\documentclass{article}

    \usepackage[numbers]{natbib}
    \usepackage[final]{neurips_2021}

\usepackage[utf8]{inputenc} 
\usepackage[T1]{fontenc}    
\usepackage{hyperref}       
\usepackage{url}            
\usepackage{booktabs}       
\usepackage{amsfonts}       
\usepackage{nicefrac}       
\usepackage{microtype}      
\usepackage{xcolor}  

\usepackage{hyperref}       
\usepackage{url}            
\usepackage{booktabs}       
\usepackage{amsfonts}       
\usepackage{nicefrac}       
\usepackage{microtype}      
\usepackage{amsmath,amssymb,latexsym}
\usepackage[capitalize]{cleveref}
\usepackage{mathtools}
\usepackage{dsfont}
\usepackage{xspace}
\usepackage[thinc]{esdiff}

\DeclareMathOperator{\Pre}{Pre}
\DeclareMathOperator{\rank}{rank}
\newcommand{\ie}{\emph{i.e.}\xspace}
\def\eg{\textit{e.g.}\xspace}
\def\vs{\textit{vs}\xspace}
\usepackage{caption}
\usepackage{subcaption}
\usepackage{pgfplots}
\usepackage{pifont}
\usepackage{array,multirow}
\usepackage{tabularx}
\newcolumntype{Y}{>{\centering\arraybackslash}X}
\usepackage{wrapfig,lipsum,booktabs}
\newcommand{\cmark}{\ding{51}}
\newcommand{\xmark}{\ding{55}}
\usepackage{siunitx}
\usepackage{stmaryrd}
\usepackage{enumerate}

\usepackage{tikz}
\newcommand*\circled[1]{\tikz[baseline=(char.base)]{
            \node[shape=circle,draw,inner sep=1.2pt] (char) {#1};}}

\usepackage{todonotes}

\title{Robust and Decomposable Average Precision for~Image Retrieval}

\author{%
  Elias Ramzi\textsuperscript{1,2} \\
  \texttt{elias.ramzi@cnam.fr} \\
   \And
  Nicolas Thome\textsuperscript{1} \\
  \texttt{nicolas.thome@cnam.fr}
   \AND
   Clément Rambour\textsuperscript{1} \\
  \texttt{clement.rambour@cnam.fr}
   \And
  Nicolas Audebert\textsuperscript{1} \\
  \texttt{nicolas.audebert@cnam.fr}
  \And
  Xavier Bitot\textsuperscript{2} \\
  \texttt{xavier.bitot@coexya.eu} \medskip\\
  \textsuperscript{1}CEDRIC, Conservatoire National des Arts et Métiers, Paris, France \\
  \textsuperscript{2}Coexya, Paris, France
}

\newcommand{\Labs}{\mathcal{L}_\text{calibr.}}
\newcommand{\LAP}{\mathcal{L}_\text{AP}}
\newcommand{\LROADMAP}{\mathcal{L}_\text{ROADMAP}}
\newcommand{\LsupAP}{\mathcal{L}_\text{SupAP}}
\newcommand{\AP}{\text{AP}}
\newcommand{\LSmoothAP}{\mathcal{L}_\text{SmoothAP}}

\begin{document}

\maketitle

\begin{abstract}

In image retrieval, standard evaluation metrics rely on score ranking, e.g. average precision (AP). In this paper, we introduce a method for robust and decomposable average precision (ROADMAP) addressing two major challenges for end-to-end training of deep neural networks with AP: non-differentiability and non-decomposability. Firstly, we propose a new differentiable approximation of the rank function, which provides an upper bound of the AP loss and ensures robust training. Secondly, we design a simple yet effective loss function to reduce the decomposability gap between the AP in the whole training set and its averaged batch approximation, for which we provide theoretical guarantees. Extensive experiments conducted on three image retrieval datasets show that ROADMAP outperforms several recent AP approximation methods and highlight the importance of our two contributions. Finally, using ROADMAP for training deep models yields very good performances, outperforming state-of-the-art results on the three datasets. Code and instructions to reproduce our results will be made publicly available at \url{https://github.com/elias-ramzi/ROADMAP}.

\end{abstract}

\section{Introduction}
\label{introduction}

The task of ‘query by example’ is a major prediction problem, which consists in learning a similarity function able to properly rank all the instances in a retrieval set
according to their relevance to the query, such that relevant items have the largest similarity. In computer vision, it drives several major applications, \eg content-based image retrieval, face recognition or person re-identification. 

Such tasks are usually evaluated with rank-based metrics, \eg Recall@k, Normalized Discounted Cumulative
Gain (NDCG), and  Average
Precision (AP). AP  is also the \textit{de facto} metric used in several vision tasks implying a large imbalance between positive and negative samples, \eg object detection.

In this paper, we address the problem of direct AP training with stochastic gradient-based optimization, \eg using deep neural networks, which poses two major challenges.

Firstly, the AP loss $\LAP = 1-\text{AP}$ is not differentiable and is thus not directly amenable to gradient-based optimization. There has been a rich literature for providing smooth and upper bound surrogate losses for $\LAP$~\cite{Yue:2007,Mcfee10metriclearning,Mohapatra_2018_CVPR,Durand19,blackbox}. More recently, smooth differentiable rank approximations have been proposed~\cite{histogram_loss,He_2018_CVPR,He_2018_DOAP,fastap,naverap,Engilberge_2019_CVPR,smoothap}, but generally lose the important $\LAP$ upper bound property.

The second important issue of AP optimization relates to its non-decomposability: $\LAP^B$  averaged over batches underestimates $\LAP$ on the whole training dataset, which we refer as the \textit{decomposability gap}. In image retrieval, the attempts to circumvent the problem involve \textit{ad hoc} methods based on batch sampling strategies~\cite{Ge_2018_ECCV,Suh_2019_CVPR,DBLP:conf/iccv/ManmathaWSK17,Suh_2019_CVPR,NIPS2016_6b180037}, or storing all training representations/scores~\cite{xbm,fastap,naverap,blackbox}, leading to complex models with a large computation and memory overhead.

In this paper, we introduce a method for RObust And DecoMposable Average Precision (ROADMAP), which explicitly addresses the aforementioned challenges of AP optimization.

Our first contribution is to propose a new surrogate loss $\LsupAP$ for $\LAP$. In particular, we introduce a smooth approximation of the rank function, with a different behaviour for positive and negative examples. By this design, $\LsupAP$ provides an upper bound of $\LAP$, and always back-propagates gradients when the correct ranking is not satisfied. These two features illustrated in the the toy example on Figure are not fulfilled by binning approaches \cite{fastap,naverap} or by SmoothAP~\cite{smoothap}.

\begin{figure*}[t]
    \centering
    \begin{subfigure}[t]{0.47\textwidth}
        \centering
        \includegraphics[scale=.4, height=3.5cm]{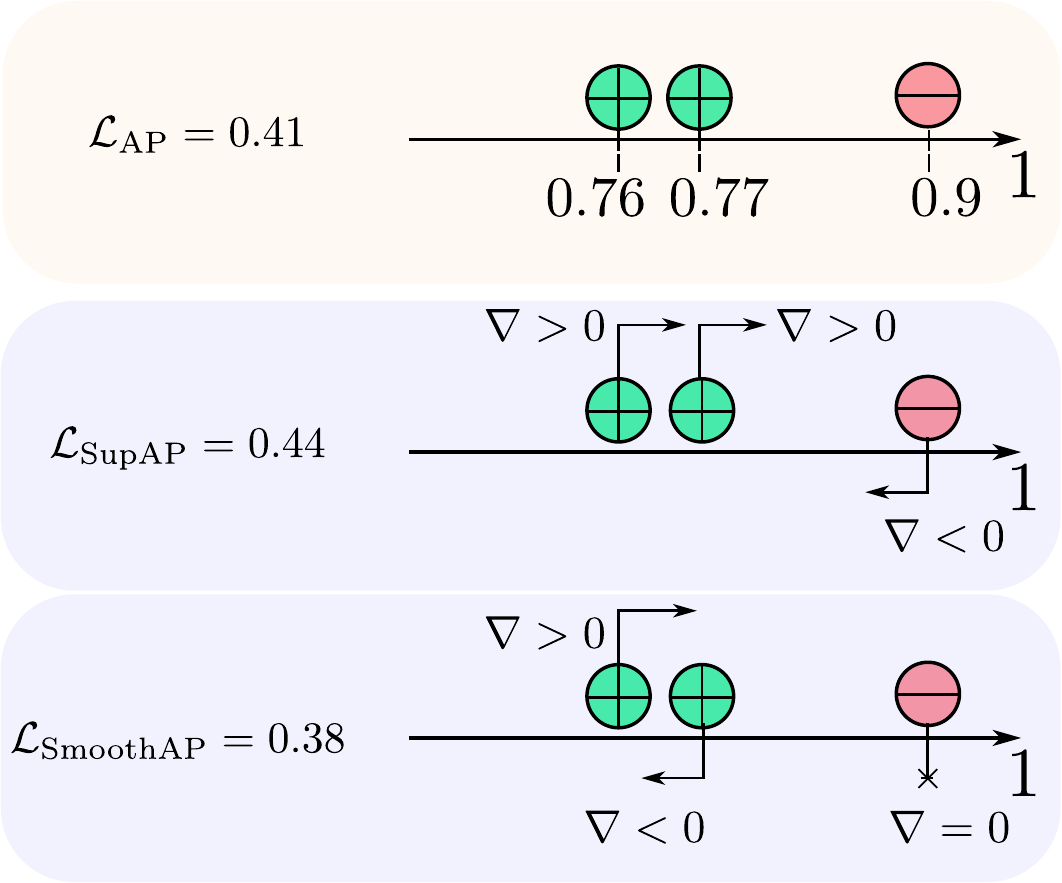}
        \caption{$\LsupAP \geq \LAP$ and $\nabla \LsupAP >0$ in this example, in contrast to SmoothAP \cite{smoothap}. This ensures robust training and  comes from a new approximation of the rank function.}
        \label{fig:introa}
    \end{subfigure}
    ~ 
    \begin{subfigure}[t]{0.47\textwidth}
        \centering
        \includegraphics[width=1\textwidth, height=3.5cm]{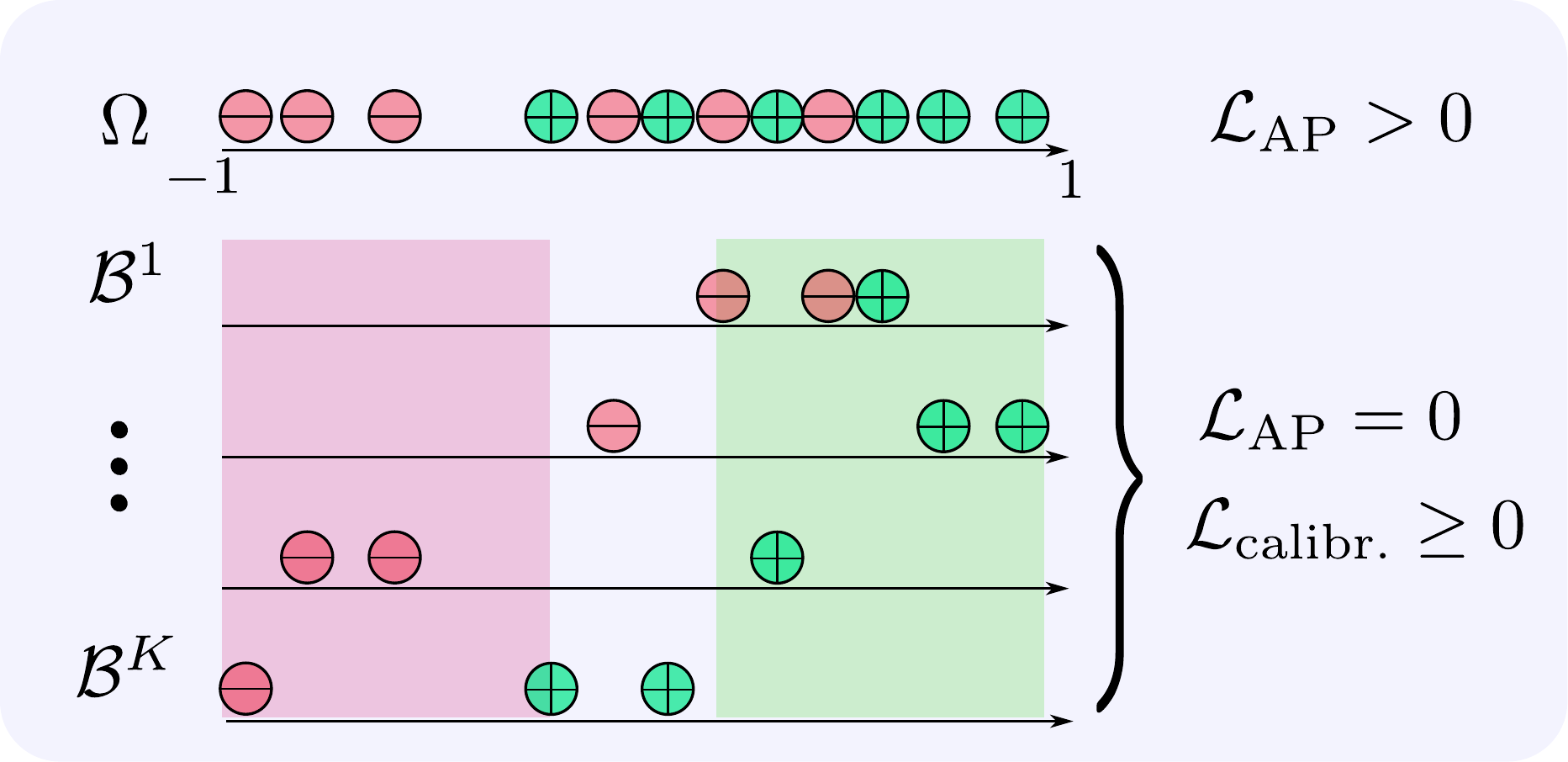}
        \caption{$\LAP$ non-decomposability: $\LAP=0$ in all batches $\mathcal{B}^i$ despite $\LAP \neq 0$ over the whole $\bigcup_i \mathcal{B}^i$. $\Labs$ controls the absolute scores between batches, such that $\LROADMAP \neq 0$ in each batch.}
        \label{fig:introb}
    \end{subfigure}
    \caption{Our robust and decomposable Average Precision training (ROADMAP) includes (a) a smooth loss $\LsupAP$ upper-bounding $\LAP$, and (b) a calibration loss $\Labs$ supporting decomposability.}
    \label{fig:intro}
\end{figure*}

As a second contribution, we propose to improve the non-decomposability in AP training. To this end, we introduce a simple yet effective training objective $\Labs$, which calibrates the scores among different batches by controlling the absolute value of positive and negative samples. We provide a theoretical analysis  showing that $\Labs$ decreases the decomposability gap. Figure \ref{fig:introb} illustrates how $\Labs$ can be leveraged to improve the overall ranking.

We provide a thorough experimental validation including three standard image retrieval datasets and show that ROADMAP outperforms state-of-the-art methods. We also report the large and consistent gain compared to rank/AP approximation baselines, and we highlight in the ablation studies the importance of our two contributions. Finally, ROADMAP does not entail any memory or computation overhead and remains competitive even with small batches.

\section{Related work}
\label{related_work}
We discuss here the literature in image retrieval dedicated to AP optimization, and compare to other approaches based on optimizing representations~\cite{proxynca,unifying_mi,norm_softmax,fewer_is_more,proxynca++} in the experiments.

\textbf{Smooth AP approximations}
Studying smooth surrogate losses for AP has a long history. The widely used surrogate for retrieval is to consider constraints based on pairs~\cite{NIPS2002_c3e4035a,hadsell2006dimensionality,DBLP:conf/eccv/RadenovicTC16}, triplets~\cite{DBLP:journals/ijcv/GordoARL17}, quadruplets~\cite{DBLP:journals/ijcv/LawTC17} or n-uplets~\cite{NIPS2016_6b180037} to enforce partial ranking. These metric learning methods
~optimize a very coarse upper bound on AP and need complex post-processing and tricks to be effective.

One option for training with AP is to design smooth upper bounds on the AP loss.
Seminal works are based on structural SVMs~\cite{Yue:2007,Mcfee10metriclearning}, with extensions to speed-up the "loss-augmented inference"~\cite{Mohapatra_2018_CVPR} or to adapt to weak supervision~\cite{Durand19}. Recently, a generic blackbox combinatorial solver has been introduced~\cite{blackbox} and applied to AP optimization~\cite{blackboxap}. To overcome the brittleness of AP with respect to small score variations, an \textit{ad hoc} perturbation is applied to positive and negative scores  during training. 
These methods provide elegant AP upper bounds, but generally are coarse AP approximations.

Other approaches rely on designing smooth approximations of the the rank function. This is done in soft-binning techniques~\cite{He_2018_CVPR,He_2018_DOAP,histogram_loss,fastap,naverap} by using a smoothed discretization of similarity scores. Other approaches rely on explicitly approximating the non-differentiable rank functions using neural networks~\cite{Engilberge_2019_CVPR}, or with a sum of sigmoid functions in the recent SmoothAP approach~\cite{smoothap}. These approaches enable accurate AP approximations by providing tight and smooth approximations of the rank function. However, they do not guarantee that the resulting loss is an AP loss upper bound. The $\LsupAP$ introduced in this work is based on a smooth approximation of the rank function leading to an upper bound on the AP loss, making our approach both accurate and robust.

\textbf{Decomposability in AP optimization} Batch training is mandatory in deep learning. However, the non-decomposability of AP is a severe issue, since it yields an inconsistent
~AP gradient estimator. 

Non-decomposability is related to sampling informative constraints in simple AP surrogates, \eg triplet losses, since the constraints' cardinality on the whole training set is prohibitive. This has been addressed by efficient batch sampling~\cite{Harwood_2017_ICCV,Ge_2018_ECCV,Suh_2019_CVPR} or selecting informative constraints within mini-batches ~\cite{NIPS2016_6b180037,VSE++,DBLP:conf/sigir/CarvalhoCPSTC18,Suh_2019_CVPR}. In cross-batch memory technique~\cite{xbm}, the authors assume a slow drift in learned representations to store them and compute global mining in pair-based deep metric learning.

In AP optimization, the non-decomposability has essentially been addressed by a brute force increase of the batch size \cite{fastap,naverap,blackbox}. This includes an important overhead in computation and memory, generally involving a two-step approach for first computing the AP loss and subsequently re-computing activations and back-propagating gradients. In contrast, our loss $\Labs$ does not add any overhead and enables good performances for AP optimization even with small batches.

\section{Robust and decomposable AP training}
\label{method}
\begin{figure}[t]
    \centering
    \includegraphics[width=1\textwidth]{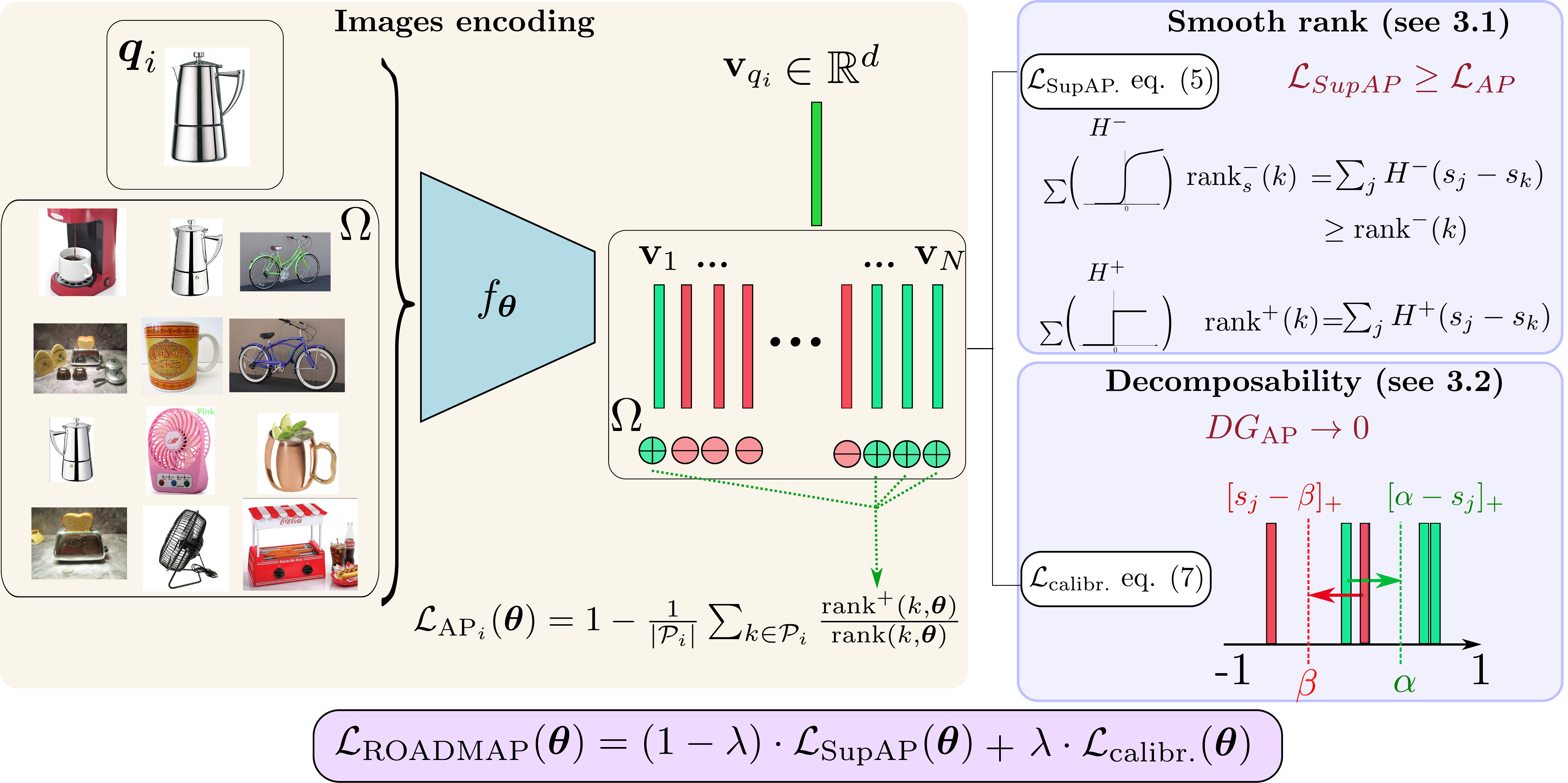}
    \caption{ROADMAP training: we optimize parameters $\boldsymbol{\theta}$ of a deep neural networks to minimize a smooth surrogate of $\mathcal L_{\AP_i}(\boldsymbol{\theta})$ between the query $\boldsymbol{q_i}$ and the retrieval set $\Omega$. Our smooth rank approximations $H^+$ and $H^-$ enables $\LsupAP$ to be both accurate and robust (sec \ref{sec:supAP}), and $\Labs$ enables an implicit batch scores comparison for better decomposability without additional storing (sec \ref{sec:decomposableAP}).}
    \label{fig:roadmap}
\end{figure}

We present here our method for RObust And DecoMposable AP (ROADMAP) dedicated to direct optimization of a smooth surrogate of AP with stochastic gradient descent (SGD), see \cref{fig:roadmap}. 

\textbf{Training context} ~~Let us consider a retrieval set $\Omega=\left\{\boldsymbol{x_j}\right\}_{j \in \llbracket 1;N\rrbracket}$ composed of $N$ elements,
~and a set of $M$ queries included in $\Omega$, \ie $\mathcal{Q}=\left\{\boldsymbol{q_i}\right\}_{i \in \llbracket 1;M\rrbracket} \subseteq \Omega$. For each query $\boldsymbol{q_i}$, each element in $\Omega$ is assigned a label $y(\boldsymbol{x_j},\boldsymbol{q_i}) \in \left\{ +1;-1 \right\}$, such that $y(\boldsymbol{x_j},\boldsymbol{q_i})= 1$ (resp. $y(\boldsymbol{x_j},\boldsymbol{q_i})=-1$) if $\boldsymbol{x_j}$ is relevant (resp. irrelevant) with respect to $\boldsymbol{q_i}$. This defines a query-dependent partitioning of $\Omega$ such that $\Omega=\mathcal{P}_{i}\cup \mathcal{N}_{i}$, where $\mathcal{P}_{i} := \left\{\boldsymbol{x_j} \in \Omega | y(\boldsymbol{x_j},\boldsymbol{q_i})=+1\right\}$ and $\mathcal{N}_{i} := \left\{\boldsymbol{x_j} \in \Omega | y(\boldsymbol{x_j},\boldsymbol{q_i})=-1\right\}$.

For each $\boldsymbol{x_j} \in \Omega$
, we define a prediction model parametrized by parameters $\boldsymbol{\theta}$, \eg a deep neural network, which provides a vectorial embedding $\mathbf{v_{\boldsymbol{q_i}}}\in \mathbb{R}^d$ of each element, \ie: $\mathbf{v_{\boldsymbol{q_i}}}:=f_{\boldsymbol{\theta}}(\boldsymbol{q_i})$. In the embedded space $\mathbb{R}^d$, we compute a similarity score between each query $\boldsymbol{q_i}$ and each element in $\Omega$, \eg by using the cosine similarity: $s(\boldsymbol{q_i},\boldsymbol{x_j}) = \frac{\mathbf{v_{\boldsymbol{q_i}}}^T \mathbf{v_{j}}}{||\mathbf{v_{q_i}}||^2 ||\mathbf{v_{j}}||^2}$.

During training, our goal is to optimize, for each query $\boldsymbol{q_i}$, the model parameters $\boldsymbol{\theta}$ such that positive elements are ranked before negatives.
More precisely, we aim at minimizing the AP loss $\mathcal{L}_{\text{AP}_i}$ for each query $\boldsymbol{q_i}$ in the retrieval set $\Omega$. 

Our overall AP loss $\LAP$ is averaged over all queries: 
\begin{equation}
  \LAP(\boldsymbol{\theta}) = 1-\frac{1}{M} \sum_{i=1}^M  \AP_i(\boldsymbol{\theta})  \text{,~~~~}  \AP_i(\boldsymbol{\theta}) = \frac{1}{|\mathcal{P}_{i}|} \sum_{k\in \mathcal{P}_{i}} \Pre(k, \theta) =
  \frac{1}{|\mathcal{P}_{i}|} \sum_{k\in \mathcal{P}_{i}} \frac{\rank^+(k, \theta)}{\rank(k,\theta)}
    \label{eq:average_precision_with_ranks}
\end{equation}
where $\Pre(k,\theta)$ is the precision for the $k$\textsuperscript{th} positive example $\boldsymbol{x_k}$
, $\rank^+(k,\theta)$ its rank among positives $\mathcal{P}_{i}$, and the $\rank(k,\theta)$ its rank over $\Omega=\mathcal{P}_{i}\cup \mathcal{N}_{i}$.

As previously mentioned, there are two main challenges with SGD optimization of AP in Eq.~(\ref{eq:average_precision_with_ranks}): i) $\AP(\boldsymbol{\theta})$ is not differentiable with respect to $\boldsymbol{\theta}$, and ii) AP does not linearly decompose into batches.
~ROADMAP addresses both issues: we introduce the robust differentiable $\LsupAP$ surrogate (\cref{sec:supAP}), and add the $\Labs$ loss (\cref{sec:decomposableAP}) to improve AP decomposability. Our final loss $\LROADMAP$ is a linear combination of $\LsupAP$ and $\Labs$, weighted by the hyperparameter $\lambda$:
\begin{equation}
\LROADMAP(\boldsymbol{\theta}) = (1-\lambda) \cdot \LsupAP(\boldsymbol{\theta}) + \lambda \cdot \Labs(\boldsymbol{\theta})
    \label{eq:roadmap}
\end{equation}

\subsection{Robustness in smooth rank approximation}
\label{sec:supAP}

The non-differentiablity in Eq~(\ref{eq:average_precision_with_ranks}) comes from the ranking operator, which can be viewed as counting the number of instances that have a similarity score greater than the considered instance, \ie\footnote{For the sake of readability we drop in the following the dependence on $\boldsymbol{\theta}$ for the rank, \ie $\rank(k):=\rank(k,\theta)$ and on the query for the similarity, \ie $s_j:=s(q_i, x_j)$.}:
\begin{align}
    & \rank^+(k) = 1 + \sum_{j\in \mathcal{P}_{i} \setminus \{k\}} H(s_j - s_k)  \text{,~~~where~} H(t) = 
    \begin{cases}
      1 \quad \text{if} \; t \geq 0 \\
      0 \quad \text{otherwise}
    \end{cases} \nonumber \\
 & \rank(k) = 
 \rank^+(k) + \sum_{j\in  \mathcal{N}_{i}} H(s_j - s_k) = \rank^+(k) + \rank^-(k) 
    \label{eq:definition_rank}
\end{align}

From \cref{eq:definition_rank} it becomes clear that the non-differentiablity is due to the Heaviside (step) function $H$, whose derivative is either zero or undefined. Note that the computation of $\rank^+(k)$ and $\rank^-(k)$ in \cref{eq:definition_rank} relates to the rank of positive instances $\boldsymbol{x_k}\in \mathcal{P}_{i}$: the score $s_k$ in \cref{eq:definition_rank} is always the score of a positive, whereas $s_j$ can either be a negative's or positive's score.

\textbf{Smooth loss $\LsupAP$}
~To provide a smooth approximation of $\LAP$ in \cref{eq:average_precision_with_ranks}, we introduce a smooth approximation of the rank function. In particular, we propose a different behaviour between $\rank^+(k)$ and $\rank^-(k)$ in \cref{eq:definition_rank} by defining two functions $H^+$ and $H^-$. 

\begin{figure*}[t]
    \centering
        
    \begin{subfigure}[t]{0.33\textwidth}
        \centering
        \includegraphics[width=1\textwidth]{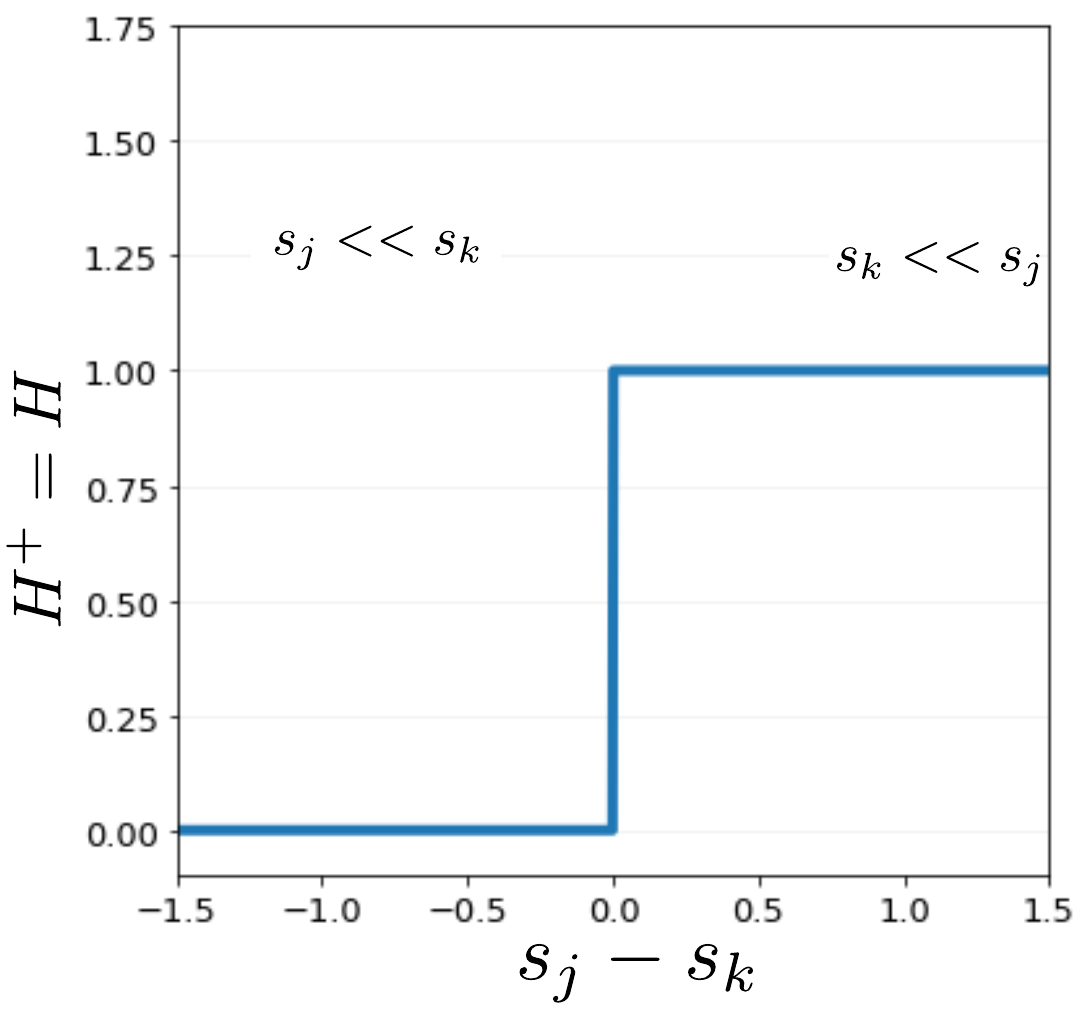}
        \caption{$H^+(x)=H(x)$ in \cref{eq:definition_rank}}
        \label{fig:h_plus}
    \end{subfigure}%
    \begin{subfigure}[t]{0.33\textwidth}
        \centering
        \includegraphics[width=1\textwidth]{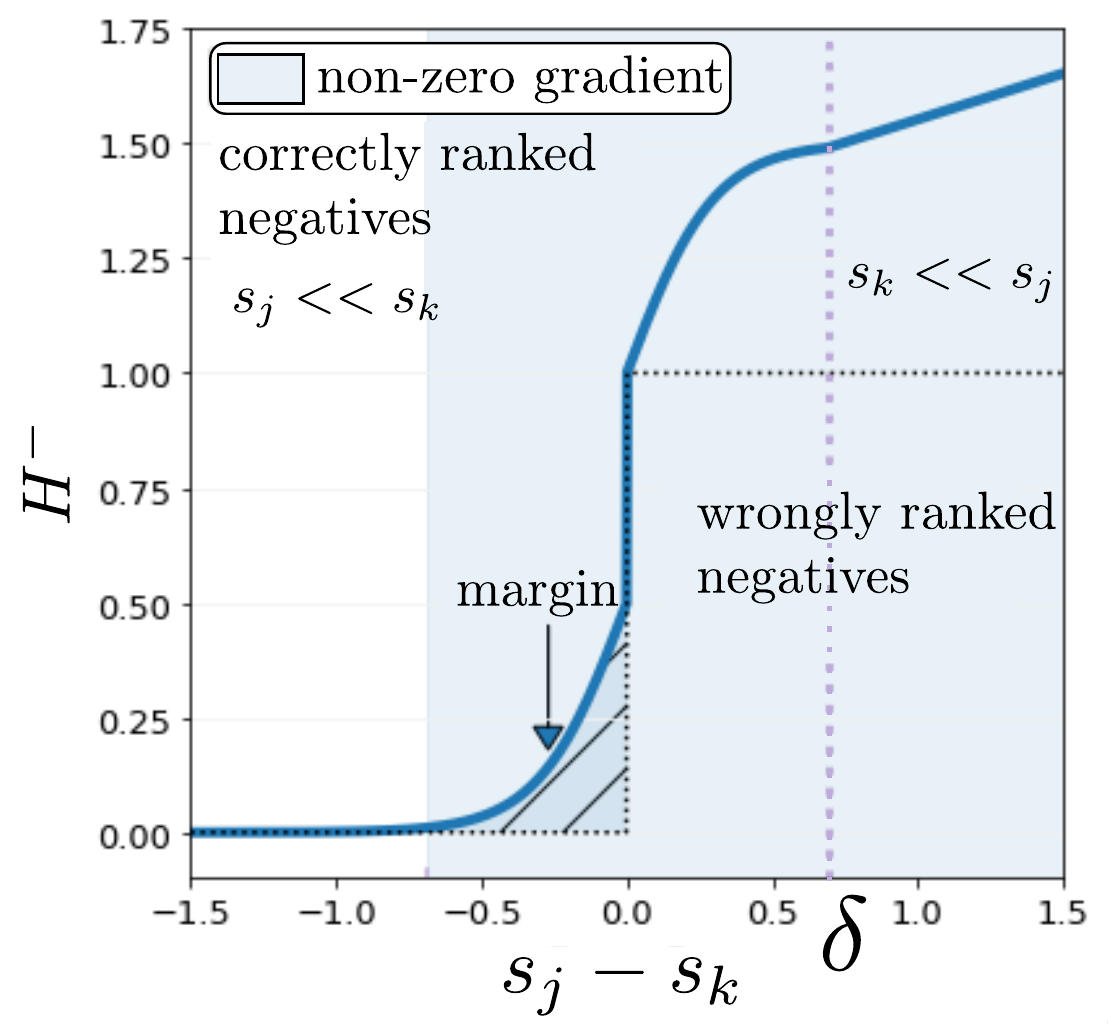}
        \caption{$H^-(x)$ in \cref{eq:h_minus}}
        \label{fig:h_minus}
    \end{subfigure}%
    \begin{subfigure}[t]{0.33\textwidth}
        \centering
        \includegraphics[width=1\textwidth]{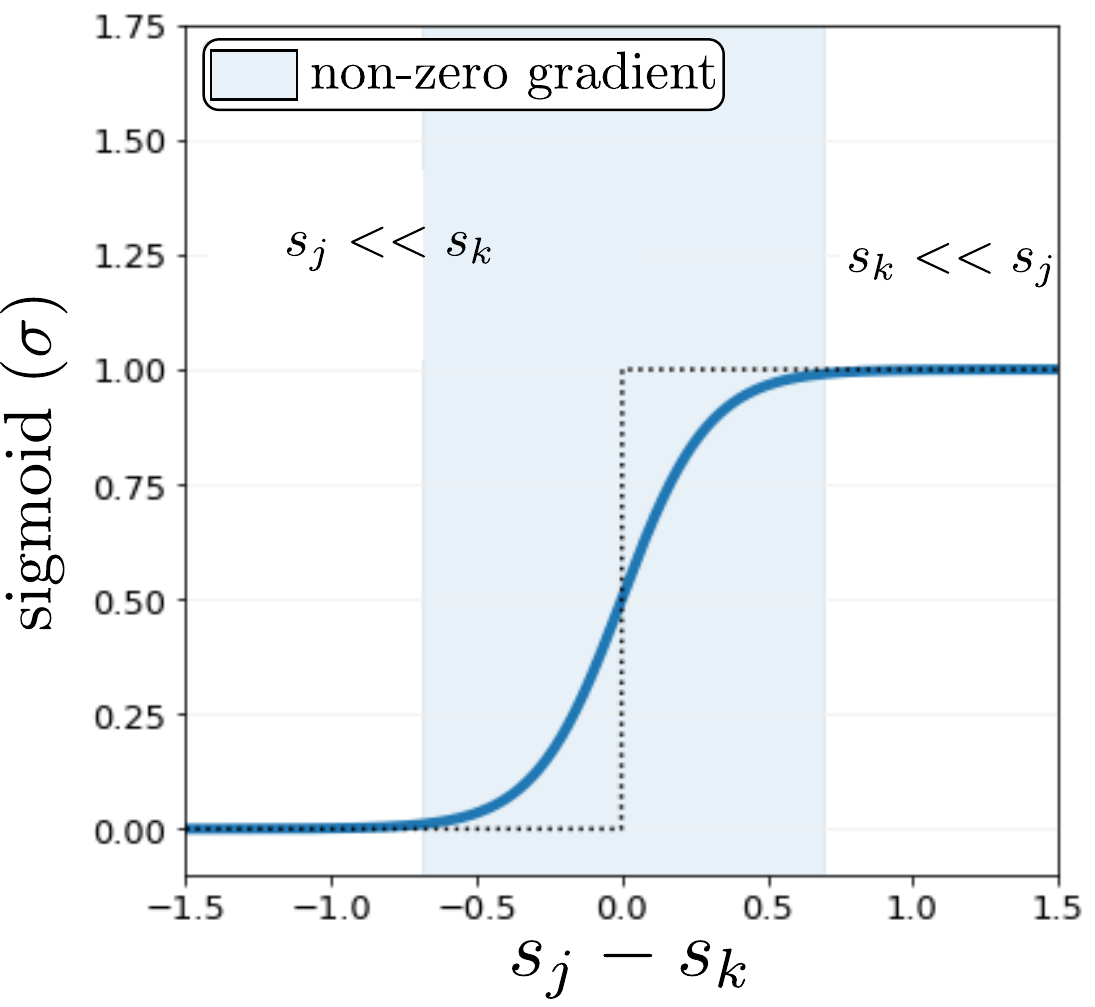}
        \caption{Sigmoid used in \cite{smoothap}}
        \label{fig:sigmoid}
    \end{subfigure}

    \caption{Proposed surrogate losses for the Heaviside (step): with $H^+(x)$ in \cref{fig:h_plus} and $H^-(x)$ in \cref{fig:h_minus}, $\LsupAP$ in \cref{eq:SmUp-AP} is an upper bound of $\LAP$. In addition, $H^-(x)$ back-propagates gradients until the correct ranking is satisfied, in contrast to the sigmoid used in \cite{smoothap} (\cref{fig:sigmoid}). }
    \label{fig:smooth_rank}
\end{figure*}

For $\rank^+(k)$, we choose to keep the Heaviside (step) function, \ie $H^+=H$ (see Fig. \ref{fig:h_plus}), which consists in ignoring $\rank^+(k)$ in gradient-based AP optimization.
~This is done on purpose since $\frac{\partial AP}{{\partial \rank^+(k)}} = \frac{\rank^-(k)}{(\rank^+(k)+\rank^-(k))^2}\geq 0$: the gradient would tend to increase $\rank^+(k)$ and to decrease the score of $s_k$.
~Reminding $\boldsymbol{x_k}$ is always a positive instance, this behaviour is undesirable.

For $\rank^-(k)$, we define the following smooth surrogate $H^-$ for $H$, shown in Fig~\ref{fig:h_minus}:

\begin{equation}
    H^-(t) = 
    \begin{cases}
      \sigma(\frac{t}{\tau}) \quad \text{if} \; t \leq 0, \quad \text{where $\sigma$ is the sigmoid function (Fig. \ref{fig:sigmoid})} \\
      \sigma(\frac{t}{\tau}) + 0.5 \quad \text{if} \; t \in [0;\delta] \quad \text{with} \; \delta \geq 0\\
      \rho \cdot (t - \delta) + \sigma(\frac{\delta}{\tau}) + 0.5 \quad \text{if} \; t > \delta \\
    \end{cases}
    \label{eq:h_minus}
\end{equation}
where $\tau$ and $\rho$ are hyperparameters, and $\delta$ is defined such that the sigmoidal part of $H^-$ reaches the saturation regime and is fixed for the rest of the paper (see supplementary Sec. A). From the $H^-$ smooth approximation defined in \cref{eq:h_minus}, we obtain the following smooth approximation $\rank_s^-(k)=\sum\limits_{j\in  \mathcal{N}_{i}} H^-(s_j - s_k) $, leading to the following smooth AP loss approximation: 
\begin{equation}
\LsupAP(\boldsymbol{\theta}) = 1-\frac{1}{M} \sum_{i=1}^M \frac{1}{|\mathcal{P}_{i}|} \sum_{k\in \mathcal{P}_{i}} \frac{\rank^+(k)}{\rank^+(k)+\rank_s^-(k)}
  \label{eq:SmUp-AP}
\end{equation}
\textbf{$\LsupAP$ in \cref{eq:SmUp-AP} fulfills two main features for AP optimization:}\vspace{0.1cm}\\ 
$~~~\blacktriangleright$ \textbf{\circled{1} $\LsupAP$ is an upper bound of $\boldsymbol{\LAP}$ in \cref{eq:average_precision_with_ranks}.} Since $H^-$ in \cref{eq:h_minus} is an upper bound of a step function (Fig~\ref{fig:h_minus}), it is easy to see that~${\mathcal{L}_{SupAP} \geq \LAP}$.
~This is a very important property, since it ensures that the model keeps training until the correct ranking is obtained. It is worth noting that existing smooth rank approximations in the literature~\cite{histogram_loss,fastap,naverap,smoothap} do not fulfill this property.\\
$~~~\blacktriangleright$ \textbf{\circled{2} $\LsupAP$ brings training gradients until the correct ranking plus a margin is fulfilled.}
When the ranking is incorrect, the negative $\boldsymbol{x_j}$ is ranked before the positive $\boldsymbol{x_k}$, thus $s_j > s_k$ and $H^-(s_j-s_k)$ in \cref{eq:h_minus} has a non-null derivative. We use a sigmoid to have a large gradient when $s_j-s_k$ is small. To overcome vanishing gradients of the sigmoid for large values $s_j - s_k$, we use a linear function ensuring constant $\rho$ derivative. When the ranking is correct ($s_j < s_k$),  we enforce robustness by imposing a margin parametrized by $\tau$ (sigmoid in \cref{eq:h_minus}). This margin overcomes the brittleness of rank losses, which vanish as soon as the ranking is correct \cite{He_2018_CVPR,fastap,blackbox}.

\textbf{Comparison to SmoothAP~\cite{smoothap}} $\LsupAP$ differs from $\mathcal{L}_\text{SmoothAP}$ in~\cite{smoothap} by i) providing an upper bound on $\LAP$, ii) improving the gradient flow (\cref{fig:h_minus} vs \cref{fig:sigmoid}), and iii) overcoming adverse effects of the sigmoid for $rank^+$, as shown in \cref{fig:introa} (and in supplementary sec. A). We experimentally verify the consistent gain brought out by $\LsupAP$ over $\mathcal{L}_\text{SmoothAP}$.

\subsection{Decomposable Average Precision}
\label{sec:decomposableAP}

In \cref{eq:average_precision_with_ranks}, AP decomposes linearly between queries $\boldsymbol{q_i}$, but $\text{AP}_i$ does not decomposes linearly between samples. We therefore focus our analysis of the non-decomposability on a single query.
~For a retrieval set $\Omega$ of $N$ elements, we consider $\{\mathcal{B}^b\}_{b\in\{1:K\}}$ batches of size B, such that $N/B=K \in \mathbb{N}$. Let $\AP_i^b(\boldsymbol{\theta})$ be the AP in batch $b$ for query $\boldsymbol{q_i}$, we define the "decomposability gap" $DG_\text{AP}$ as follows: 
\begin{equation}
    \label{eq:decomposability-gap}
    DG_\text{AP}(\boldsymbol{\theta}) =  \frac{1}{K}  \sum_{b=1}^K \AP_i^b(\boldsymbol{\theta}) - \AP_i(\boldsymbol{\theta})
\end{equation}
$DG_\text{AP} $ in \cref{eq:decomposability-gap} is a direct measure of the non-decomposability of AP {(see supplementary Sec. A)}. Our motivation here is to decrease $DG_\text{AP}$, \ie to have the average AP over the batches as close as possible to the AP computed over the whole training set. To this aim, we introduce the following loss during training: 
\begin{equation}
    \label{eq:labs}
    \begin{aligned}
    \Labs(\boldsymbol{\theta}) & = \frac{1}{M} \sum_{i=1}^M \underbrace{\frac{1}{|\mathcal{P}_{i}|}
     \sum_{\boldsymbol{x_j} \in \mathcal{P}_{i}} [\alpha - s_j]_+}_{\Labs^+} + \underbrace{\frac{1}{|\mathcal{N}_{i}|} \sum_{\boldsymbol{x_j} \in \mathcal{N}_{i}} [s_j - \beta]_+}_{\Labs^-}
    \end{aligned}
\end{equation}
where $[x]_+ = \max(0,x)$. The loss ${\Labs^+}$ enforces the score of the positive $\boldsymbol{x_i} \in \mathcal{P}_{i}$ to be larger than $\alpha$, and ${\Labs^-}$ enforces the score of the negative $\boldsymbol{x_j} \in \mathcal{N}_{i}$ to be smaller than $\beta < \alpha$. $\Labs$ is a standard pair-based loss~\cite{hadsell2006dimensionality}, which we revisit in our context to "calibrate" the values of the scores between mini-batches: intuitively, the fact that the positive (resp. negative) scores are above (resp. below) a threshold in the mini-batches makes the average AP closer to the AP on the whole dataset. 

\textbf{Upper bound on the decomposabilty gap} To formalize this idea, we provide a theoretical analysis of the impact on the global ranking of $\Labs$ in \cref{eq:labs}. 
Firstly, we can see that if ${\Labs^-}={\Labs^+}=0$, on each batch, the overall AP and the AP in batches is null, \ie $DG_\text{AP}(\boldsymbol{\theta}) = 0$  and we get a decomposable AP. In a more general setting, we show that minimizing $\Labs$ on each batch reduces the decomposability gap, hence improving the decomposability of the AP.

Let's consider $K$ batches $\{\mathcal{B}^b\}_{b\in\{1:K\}}$ of batch size $B$ divided in $\mathcal{P}_{i}^{b}$ positive instances and $\mathcal{N}_{i}^{b}$ negative instances w.r.t. the query $\boldsymbol{q_i}$.
To give some insight we assume that the AP of each batch is one (\ie $ AP^b_i = 1$), and give the following upper bound of $DG_\text{AP}$ :
\begin{equation}
    \label{eq:upper_bound}
    0 \leq DG_\text{AP} \leq 
1 - \frac{1}{\sum_{b=1}^K |\mathcal{P}_i^b|}\left( \sum_{b=1}^K \sum_{j=1}^B \frac{j + |\mathcal{P}_i^1| + \dots + |\mathcal{P}_i^{b-1}|}{j + |\mathcal{P}_i^1| + \dots + |\mathcal{P}_i^{b-1}| + |\mathcal{N}_i^1| + \dots + |\mathcal{N}_i^{b-1}|} \right ) 
\end{equation}

This upper bound of the decomposability gap is given in the worst case for the global AP : the global ranking is built from the juxtaposition of the batches {(see supplementary Sec. A)}.

We can refine this upper bound by introducing the calibration loss $\Labs$ and constraining the scores of positive and negative instances to be well calibrated. On each batch we define the following quantities $E^-_b = \sum_{j \in \mathcal{N}_i^-} \mathds{1}(s_j > \beta)$ which are the negative instances that do not respect the constraints and $G^-_b = \sum_{j \in \mathcal{N}_i^-} \mathds{1}(s_j \leq \beta)$ the negative instances that do. We similarly define $E^+_b$ and $G^+_b$.
We then have the following upper bound on the decomposability gap : 

\begin{align}
    \label{eq:upped_bound_refined}
    0 \leq DG_\text{AP}  \leq  1 - \frac{1}{\sum_{b=1}^K |\mathcal{P}_i^b|} \Bigg( & \sum_{b=1}^K \bigg[ \sum_{j=1}^{G^+_b} \frac{j + G^+_1 + \dots + G^+_{b-1}}{j + G^+_1 + \dots + G^+_{b-1} + E^-_1 + \dots E^-_{b-1}} + \\
    & \sum_{j=1}^{E^+_b} \frac{j + G^+_{b} + |\mathcal{P}_i^1| + \dots + |\mathcal{P}_i^{b-1}|}{j + G^+_{b} + |\mathcal{P}_i^1| + \dots + |\mathcal{P}_i^{b-1}| + |\mathcal{N}_i^1| + \dots + |\mathcal{N}_i^{b-1}|} \bigg]  \Bigg) \nonumber
\end{align}

This refined upper bound is tighter than the upper bound of \cref{eq:upper_bound}. Our new $\Labs$ loss directly optimizes this upper bound (by explicitly optimizing $E^-_b, E^+_b, E^+_{b}, G^+_{b}$), making it tighter, hence improving the decomposability of the AP (see supplementary Sec. A).

\section{Experiments}
\label{sec:experiments}

\paragraph{Experimental setup} We evaluate ROADMAP on the following three  image retrieval datasets:\\
\textbf{CUB-200-2011}~\cite{CUB} contains \num{11788} images of birds classified into \num{200} fine-grained classes. We follow the standard protocol and use the first (resp. last) \num{100} classes for training (resp. evaluation).\\
\textbf{Stanford Online Product (SOP)} \cite{SOP} is a dataset with \num{120053} images of \num{22634} objects classified into \num{12} categories (\eg bikes, coffee makers). We use the reference train and test splits from~\cite{SOP}. \\
\textbf{INaturalist-2018}~\cite{inaturalist} is a large scale dataset of \num{461939} wildlife animals images classified into \num{8142} classes. We use the splits from \cite{smoothap} with 70\% of the classes in the train set and the rest in the test set.

\textbf{ROADMAP settings} For all experiments in \cref{sec:validation} and \cref{sec:expes-sota}, we use $\lambda=0.5$ for $\LROADMAP$ in \cref{eq:roadmap}, $\tau=0.01$ and $\rho=100$ for $\LsupAP$ in \cref{eq:SmUp-AP}, $\alpha=0.9$ and $\beta=0.6$ for $\Labs$ in \cref{eq:labs}. We study more in depth the impact of those parameters in  \cref{sec:expes-analysis}. Deep models are trained using Adam \cite{adam} for ResNet-50 backbones and AdamW \cite{adamw} for DeiT transformers \cite{deit}.\\
\textbf{Test protocol} Methods are evaluated using the standard recall at k (R@k) and mean average precision at R \cite{musgrave2020metric} (mAP@R) metrics (see supplementary Sec. B).

\subsection{ROADMAP validation}
\label{sec:validation}

In this section, all models are trained in the same setting (ResNet-50 backbone, embedding size 512, batch size 64). The comparisons thus directly measures the impact of the training loss.

\textbf{Comparison to AP approximations.}
~In \cref{tab:compa_ranking_losses}, we compare ROADMAP on the three datasets to recent AP loss approximations including the soft-binning approaches FastAP~\cite{fastap} and SoftBinAP~\cite{naverap}, the generic solver BlackBox~\cite{blackboxap}, and the smooth rank approximation~\cite{smoothap}. We use the publicly available PyTorch implementations of all these baselines.  We can see that ROADMAP outperforms all the current AP approximations by a large margin. The gain is especially pronounced on the large scale dataset INaturalist. This highlights the importance our two contributions, \ie our robust smooth AP upper bound and our AP decomposability improvement (see supplementary Sec. B).

\newcommand{\raone}{\scriptsize{R@1}}
\newcommand{\mapr}{\scriptsize{mAP@R}}
\begin{table}[h!]
    \caption{Comparison between ROADMAP and state-of-the-art AP ranking based methods.}
    \label{tab:compa_ranking_losses} 
    \centering
    \begin{tabular}{ l cc cc cc }
        \toprule
         & \multicolumn{2}{c}{CUB} & \multicolumn{2}{c}{SOP} & \multicolumn{2}{c}{INaturalist} \\
         \midrule
         Method & R@1 & mAP@R & R@1 & mAP@R & R@1 & mAP@R \\
         \midrule
         FastAP \footnotesize\cite{fastap} & 58.9 & 22.9 & 78.2 & 51.3 & 53.5 & 19.6 \\
         SoftBin \footnotesize\cite{naverap} & 61.2 & 24.0 & 80.1 & 53.5 & 56.6 & 20.1 \\
         BlackBox \footnotesize\cite{blackboxap} & 62.6 & 23.9 & 80.0 & 53.1 & 52.3 & 15.2 \\
         SmoothAP \footnotesize\cite{smoothap} & 62.1 & 23.9 & 80.9 & 54.6 & 59.8 & 20.7 \\
         \midrule
         ROADMAP & \textbf{64.2} & \textbf{25.3} & \textbf{82.0} & \textbf{56.5} & \textbf{64.5} & \textbf{25.1} \\
         \bottomrule
    \end{tabular}
\end{table}

\textbf{Comparison to memory methods.}

XBM stores the embeddings of previously seen batches to alleviate complex batch sampling and better approximate AP on the whole dataset.
Although XBM has a low memory overhead (a few hundreds megabytes on SOP), it is time consuming. We ran experiments storing the entire dataset for SOP (60k embeddings), but for INaturalist we could not train while storing all the dataset in tractable time. We chose to store the same amount of embeddings as for SOP : 60k embeddings which is about 17\% of the training set.

We can see in \cref{tab:compa_xbm} that XBM is approximately 3 times longer to train than ROADMAP. This becomes critical on INaturalist, where training  while storing 60k images takes about 3 days, and reaches only a R@1 of $60$. Consequently, ROADMAP outperforms XBM on both datasets; there is a $\sim$+2pt increase on both metrics for SOP and an especially large gap on INaturalist. In the latter, not being able to store all the embeddings affects drastically the performances of the XBM in a negative way. There is a 5pt difference in R@1 and more than 6pt in mAP@R. This demonstrates the suitability of ROADMAP on large-scale settings.

\begin{table}[h!]
    \centering
    \caption{Our method compared to cross batch memory \cite{xbm}. The unit of time is m/e which stands for minutes per epoch.}
    \label{tab:compa_xbm}
    \begin{tabular}{ l ccc ccc }
        \toprule
         & \multicolumn{3}{c}{SOP} & \multicolumn{3}{c}{INaturalist}\\
         \midrule
         Method & R@1 & mAP@R & time$\downarrow$ & R@1 & mAP@R & time$\downarrow$ \\
         \midrule
         XBM \cite{xbm} & 80.6 & 54.9 & 6 & 59.3 & 18.5 & 34  \\
         ROADMAP (ours) & \textbf{82.0} & \textbf{56.5} & \textbf{2} & \textbf{64.5} & \textbf{25.1} & \textbf{12} \\
         \bottomrule
    \end{tabular}
\end{table}

\textbf{Ablation study.}
~To study more in depth the impact of our contributions, we perform ablation studies in \cref{tab:ablation_study}. We show the improvement against SmoothAP~\cite{smoothap} when changing the sigmoid by $H^+$ and $H^-$ for $\LsupAP$ in \cref{eq:SmUp-AP}, and the use of $\Labs$ in \cref{eq:labs}. We can see that $\LsupAP$ consistently improves performances over $\mathcal{L}_\text{SmoothAP}$ (0.9pt on CUB, 0.5pt on SOP and 1.5pt on INaturalist). $\LsupAP$ and $\Labs$ equally contribute to the overall gain in CUB and SOP, but the gain of $\Labs$ is much more important on INaturalist. This is explained by the fact that the batch vs. dataset ratio size $\frac{B}{N}$ is tiny ($\ll 1$), making the decomposability gap in \cref{eq:decomposability-gap} huge. We can see that $\Labs$ is very effective for reducing this gap and brings a gain of more than 3pt. 

\begin{table}[h!]
    \vspace{-0.8\intextsep}
    \caption{Ablation study for the impact of our two contribution on and the SmoothAP baseline.}
    \label{tab:ablation_study} 
    \centering
    \begin{tabular}{ l cc cc cc cc }
        \toprule
         & & & \multicolumn{2}{c}{CUB} & \multicolumn{2}{c}{SOP} & \multicolumn{2}{c}{INaturalist}\\
         \midrule
         Method & $H^-$ & $\Labs$ & R@1 & mAP@R & R@1 & mAP@R & R@1 & mAP@R \\
         \midrule
         SmoothAP~\cite{smoothap} &  \xmark & \xmark & 62.1 & 23.9 & 80.9 & 54.6 & 59.7 & 20.7 \\
         SupAP & \cmark &  \xmark  & 62.9 & 24.6 & 81.4 & 55.3 & 61.2 & 21.3 \\
         ROADMAP &  \cmark & \cmark & \textbf{64.2} & \textbf{25.3} & \textbf{82.0} & \textbf{56.5} & \textbf{64.5} & \textbf{25.1} \\
         \bottomrule
    \end{tabular}
\end{table}

\subsection{State of the art comparison}
\label{sec:expes-sota}
We compare ROADMAP to other state of the art methods across three image retrieval datasets and report the results in~\cref{general_results_sop}. We divide competitor methods into three categories: metric learning \cite{mic,multi_similarity,spherical_embedding,horde,xbm,xuan2020hard}, classification losses for image retrieval \cite{fewer_is_more,norm_softmax,unifying_mi,proxynca++}, and AP approximations \cite{fastap,blackboxap,smoothap}. ROADMAP falls in the latter category. We use the same setup as in~\cref{sec:validation} and follow standard practices for ResNet-50~\cite{proxynca++,xuan2020hard,unifying_mi} by using larger images ($256\times256$ on SOP and CUB) and using max instead of average pooling and layer normalization for CUB.

\begin{table}[t]
    \caption{Comparison of state of the art performances from the literature on SOP, CUB and INaturalist with the proposed ROADMAP (recall@k). Except for the DeiT category, all methods rely on a standard convolutional backbone (generally ResNet-50).}
    \setlength\tabcolsep{3pt}
    \label{general_results_sop}
    \begin{tabularx}{\textwidth}{ l l>{\small}c |YYY|YYYY|YYYY }
        \toprule
        & & & \multicolumn{3}{c}{SOP} & \multicolumn{4}{c}{CUB} & \multicolumn{4}{c}{INaturalist}\\
        & Method & dim & 1 & 10 & 100 & 1 & 2 & 4 & 8 & 1 & 4 & 16 & 32\\
        \midrule
        \multirow{8}{*}{\rotatebox[origin=c]{90}{Metric learning}}
        & Triplet SH \cite{sampling_matters} & 512 & 72.7 & 86.2 & 93.8 & 63.6 & 74.4 & 83.1 & 90.0 & 58.1 & 75.5 & 86.8 & 90.7\\
        & LiftedStruct \cite{SOP} & 512 & 62.1 & 79.8 & 91.3 & 47.2 & 58.9 & 70.2 & 80.2 & -&-&-&-\\
        & MIC \cite{mic} & 512 & 77.2 & 89.4 & 95.6 & 66.1 & 76.8 & 85.6 & - &-&-&-&- \\
        & MS \cite{multi_similarity} & 512 & 78.2 & 90.5 & 96.0 & 65.7 & 77.0 & 86.3 & 91.2 &-&-&-&-\\
        & SEC \cite{spherical_embedding} & 512 & 78.7 & 90.8 & 96.6 & 68.8 & 79.4 & 87.2 & 92.5 &-&-&-&-\\
        & HORDE \cite{horde} & 512 & 80.1 & 91.3 & 96.2 & 66.8 & 77.4 & 85.1 & 91.0 &-&-&-&-\\
        & XBM \cite{xbm} & 128 & 80.6 & 91.6 & 96.2 & 65.8 & 75.9 & 84.0 & 89.9 &-&-&-&-\\
        & Triplet SCT \cite{xuan2020hard} & 512/64 & 81.9 & 92.6 & 96.8 & 57.7 & 69.8 & 79.6 & 87.0 &-&-&-&-\\
        \midrule
        \multirow{7}{*}{\rotatebox[origin=c]{90}{Classification}}
        & ProxyNCA \cite{proxynca} & 512 & 73.7 & - & - & 49.2 & 61.9 & 67.9 & 72.4 & 61.6 & 77.4 & 87.0 & 90.6 \\
        & ProxyGML \cite{fewer_is_more} & 512 & 78.0 & 90.6 & 96.2 & 66.6 & 77.6 & 86.4 & - &-&-&-&-\\
        & NSoftmax \cite{norm_softmax} & 512 & 78.2 & 90.6 & 96.2 & 61.3 & 73.9 & 83.5 & 90.0 &-&-&-&-\\
        & NSoftmax \cite{norm_softmax} & 2048 & 79.5 & 91.5 & 96.7 & 65.3 & 76.7 & 85.4 & 91.8 &-&-&-&-\\
        & Cross-Entropy \cite{unifying_mi} & 2048 & 81.1 & 91.7 & 96.3 & 69.2 & 79.2 & 86.9 & 91.6 &-&-&-&-\\
        & ProxyNCA++ \cite{proxynca++} & 512 & 80.7 & 92.0 & 96.7 & 69.0 & 79.8 & 87.3 & 92.7 &-&-&-&-\\
        & ProxyNCA++ \cite{proxynca++} & 2048 & 81.4 & 92.4 & 96.9 & 72.2 & 82.0 & 89.2 & 93.5 &-&-&-&-\\
        \midrule
        \multirow{6}{*}{\rotatebox[origin=c]{90}{AP loss}}
        & FastAP \cite{fastap} & 512 & 76.4 & 89.0 & 95.1 & -&-&-&- & 60.6 & 77.0 & 87.2 & 90.6\\
        & BlackBox \cite{blackboxap} & 512 & 78.6 & 90.5 & 96.0 & 64.0 & 75.3 & 84.1 & 90.6 & 62.9 & 79.4 & 88.7 & 91.7\\
        & SmoothAP \cite{smoothap} & 512 & 80.1 & 91.5 & 96.6 &-&-&-&- & 67.2 & 81.8 & 90.3 & 93.1\\ 
        & SoftBin\textsuperscript{*} \cite{naverap} & 512 & 80.6 & 91.3 & 96.1 & 61.2 & 73.14 & 83.0 & 89.5 & 64.2 & 77.1 & 82.7 & 91.7 \\
        & ROADMAP (ours) & 512 & 83.1 & 92.7 & 96.3 & 68.5 & 78.7 & 86.6 & 91.9 & 69.1 & 83.1 & 91.3 & 93.9\\
        \midrule
        \multirow{2}{*}{\rotatebox[origin=c]{90}{DeiT}}
        & IRT\textsubscript{R} \cite{transformer_ir} & 384 & 84.2 & 93.7 & 97.3 & 76.6 & 85.0 & 91.1 & 94.3 &-&-&-&-\\
        & ROADMAP (ours) & 384 & \textbf{86.0} & \textbf{94.4} & \textbf{97.6} & \textbf{77.4} & \textbf{85.5} & \textbf{91.4} & \textbf{95.0} & \textbf{73.6} & \textbf{86.2} & \textbf{93.1} & \textbf{95.2} \\
        \bottomrule
    \end{tabularx}
\end{table}

Using the popular ResNet-50 backbone, ROADMAP establishes a new state of the art across all methods for SOP and the challenging INaturalist dataset and outperforms all previous AP approximations on CUB, while being competitive with the other two top performers (ProxyNCA++ and SEC). R@k improvements are consistent on all datasets with a $\sim$2pts R@1 increase on INaturalist and $\sim$3pts increase on SOP compared to SmoothAP, the best performing AP approximation from the literature.

    Switching the backbone to the more recent vision transformer architecture DeiT~\cite{vit,deit}, further lifts the performances of ROADMAP by several point,
    ~from 3 to 9 points depending on the dataset, with a smaller embedding size (384 \vs 512).
    ~The decomposable AP approximation ROADMAP also outperforms by a significant margin IRT\textsubscript{R}, the DeiT architecture for image retrieval introduced in \cite{transformer_ir} trained with a contrastive loss. 
    ~Overall ROADMAP achieves state-of-the-art performances across all three datasets by a significant margin.

\subsection{Model Analysis}
\label{sec:expes-analysis}

\begin{figure*}[h!]
    \centering
    \begin{subfigure}[t]{0.32\textwidth}
        \begin{tikzpicture}[scale=0.5]
        \begin{axis}[
            title={},
            xlabel={},
            ylabel={mAP@R},
            xmin=0, xmax=1,
            ymin=22.5, ymax=28,
            xtick={0,0.2,0.5,0.8,1.0},
            ytick={},
            legend pos=south east,
            ymajorgrids=true,
            grid style=dashed,
            font=\LARGE,
        ]
        \addplot[
            line width=2,
            color=blue,
            mark=x,
            mark size=5,
            ]
            coordinates {
            (0.0,26.13)(0.2,27.64)(0.5,27.74)(0.8,26.9)(1.0,23.78)
            };
        \end{axis}
        \end{tikzpicture}
      \caption{mAP@R \vs $\lambda$ for $\LROADMAP$}
      \label{fig:lambda_roadmap}
    \end{subfigure}
    ~
    \begin{subfigure}[t]{0.32\textwidth}
        \begin{tikzpicture}[scale=0.5]
        \begin{axis}[
            title={},
            xlabel={},
            ylabel={},
            xmin=0.03, xmax=10000,
            ymin=24, ymax=26.5,
            xtick={0.1,1.0,10.0,100,1000,10000},
            ytick={},
            legend pos=south east,
            ymajorgrids=true,
            grid style=dashed,
            xmode=log,
            font=\LARGE,
        ]
        \addplot[
            line width=2,
            color=blue,
            mark=x,
            mark size=5,
            ]
            coordinates {
            (0.03,26.0)(0.1,26.02)(1.0,26.13)(10.0,26.42)(100,26.32)(1000,24.96)(10000,24.3)
            };
        \end{axis}
        \end{tikzpicture}
        \caption{mAP@R \vs $\rho$ for $\LsupAP$}
        \label{fig:rho_supap}
    \end{subfigure}
    ~
    \begin{subfigure}[t]{0.32\textwidth}
        \begin{tikzpicture}[scale=0.5]
        \begin{axis}[
            title={},
            xlabel={},
            ylabel={},
            xmin=0, xmax=1.0,
            ymin=15, ymax=28.5,
            xtick={0,0.2,0.4,0.6,0.8},
            ytick={},
            legend pos=south west,
            ymajorgrids=true,
            grid style=dashed,
            font=\LARGE,
        ]
        \addplot[
            line width=2,
            color=blue,
            mark=x,
            mark size=5,
            ]
            coordinates {
            (0.1,27.57)(0.2,27.75)(0.3,27.74)(0.4,27.68)(0.5,27.11)(0.6,26.47)(0.7,24.0)(0.8,19.41)(0.9,15.52)
            };
            
        \addplot[
        line width=1.6,
        dashed,
        color=red,
        ]
        coordinates {
        (0.0,26.13)(1.0,26.13)
        };
        \legend{$\LROADMAP$,$\LsupAP$}
        \end{axis}
        \end{tikzpicture}
      \caption{mAP@R \vs $\alpha-\beta$ for $\Labs$}
      \label{fig:margin_roadmap}
    \end{subfigure}

    \caption{Analysis of ROADMAP hyperparameters on INaturalist (batch size 224).}
    \label{fig:supap_hyperparameters}
\end{figure*}
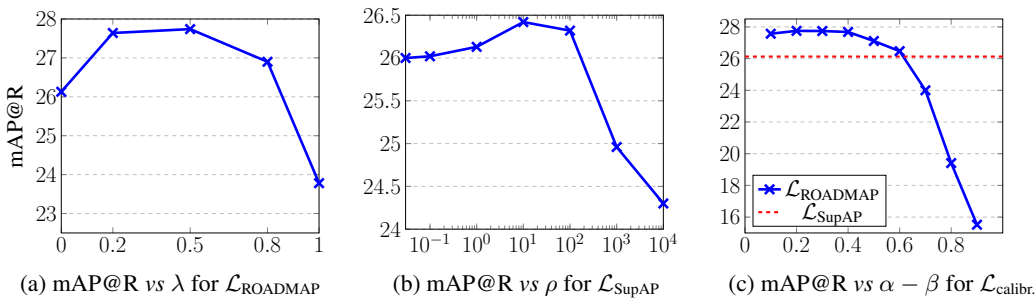

We show in \cref{fig:supap_hyperparameters} the impact of the main ROADMAP hyperparameters on INaturalist. The relative weighting $\lambda$ from \cref{eq:roadmap} controls the balance between our two training objectives $\LsupAP$ and $\Labs$: $\lambda = 0$ reduces $\LROADMAP$ to $\LsupAP$ while $\lambda=1$ to $\Labs$. We can see in~\cref{fig:lambda_roadmap} that training with the complete $\LROADMAP$ with both $\Labs$ and $\LsupAP$ is always better than using only one of the two losses. Note that results are stable in the $[0.2, 0.8]$ range with a consistent $\sim$1.5pt increase, demonstrating the robustness of ROADMAP to this hyperparameter tuning.\\

\cref{fig:rho_supap} shows the influence of the slope $\rho$ that controls the linear regime in $H^-$ and determines the amount of gradient backpropagated for negative samples with a (wrong) high score. As shown in \cref{fig:rho_supap}, the improvement is important and stable in $[10, 100]$. Note that $\rho>0$ already improves the results compared to $\rho=0$ in \cite{smoothap}. There is an important decrease when $\rho \gg 100$ probably due to the high gradient that takes over the signal for correctly ranked samples.\\
The impact of the margin $\alpha - \beta$ in $\Labs$ is shown in \cref{fig:margin_roadmap}. Once again, ROADMAP exhibits a robust behaviour w.r.t. the values of its hyperparameters: any margin in the $[0.1, 0.6]$ range results in an improvement in mAP@R compared to the $\LsupAP$ baseline without the decomposability loss. Best results are achieved with smaller margins $0.1 < \alpha - \beta < 0.4$.

\cref{fig:relative_increase} shows the improvement in mAP@R on the three datasets when adding $\Labs$ to $\LsupAP$. We can see that the increase becomes larger as the batch size gets smaller.
This confirms our intuition that the decomposability in $\Labs$ has a stronger effect on smaller batch sizes, for which the AP estimation is noisier and $DG_\text{AP}$ larger. This is critical on the large-scale dataset INaturalist where the batch AP on usual batch sizes is a very poor approximation of the global AP.

\definecolor{redsop}{RGB}{250,164,146}
\definecolor{greeninat}{RGB}{198,250,146}
\definecolor{bluecub}{RGB}{146,225,250}

\pgfplotstableread[row sep=\\,col sep=&]{
    interval & cub & sop & inat \\
    32     & 3.4 & 2.4 & 18.4 \\
    64     & 2.6 & 2.4 & 16.3  \\
    128    & 2.4 & 2.2 & 10.3 \\
    224    & 2.3 & 1.7 & 5.8 \\
    384    & 2.0 & 1.2 & 3.2 \\
    }\mydata

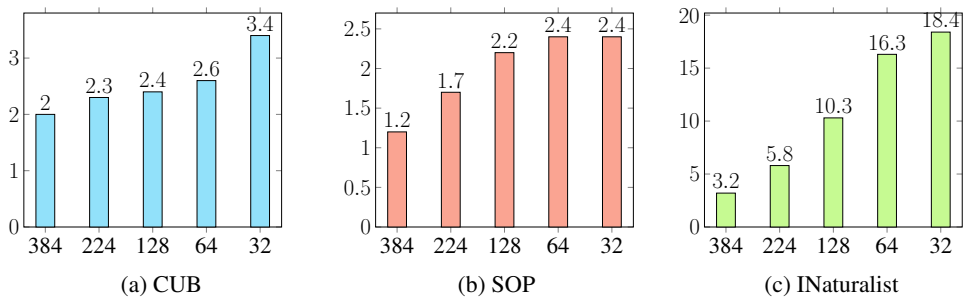
\begin{figure*}[ht]
    \centering
    \begin{subfigure}[t]{0.3\textwidth}
    \begin{tikzpicture}[scale=0.5]
        \begin{axis}[
                ybar,
                font=\LARGE,
                bar width=.5cm,
                legend style={at={(0.5,1)},
                    anchor=north,legend columns=-1},
                symbolic x coords={32,64,128,224,384},
                xtick=data,
                nodes near coords,
                nodes near coords align={vertical},
                ymin=0,ymax=3.8,
                ylabel={},
                x dir=reverse
            ]
            \addplot[fill=bluecub] table[x=interval,y=cub]{\mydata};
            \legend{}
        \end{axis}
    \end{tikzpicture}
    \caption{CUB}
    \label{fig:relative_increase_cub}
    \end{subfigure}
    ~
    \begin{subfigure}[t]{0.3\textwidth}
    \begin{tikzpicture}[scale=0.5]
        \begin{axis}[
                ybar,
                font=\LARGE,
                bar width=.5cm,
                legend style={at={(0.5,1)},
                    anchor=north,legend columns=-1},
                symbolic x coords={32,64,128,224,384},
                xtick=data,
                nodes near coords,
                nodes near coords align={vertical},
                ymin=0,ymax=2.7,
                ylabel={},
                x dir=reverse
            ]
            \addplot[fill=redsop] table[x=interval,y=sop]{\mydata};
            \legend{}
        \end{axis}
    \end{tikzpicture}
    \caption{SOP}
    \label{fig:relative_increase_sop}
    \end{subfigure}
    ~
    \begin{subfigure}[t]{0.3\textwidth}
    \begin{tikzpicture}[scale=0.5]
        \begin{axis}[
                ybar,
                font=\LARGE,
                bar width=.5cm,
                legend style={at={(0.5,1)},
                    anchor=north,legend columns=-1},
                symbolic x coords={32,64,128,224,384},
                xtick=data,
                nodes near coords,
                nodes near coords align={vertical},
                ymin=0,ymax=20.2,
                ylabel={},
                x dir=reverse
            ]
            \addplot[fill=greeninat] table[x=interval,y=inat]{\mydata};
        \end{axis}
    \end{tikzpicture}
    \caption{INaturalist}
    \label{fig:relative_increase_inat}
    \end{subfigure}
    
    \caption{Relative increase of the mAP@R \vs batch size when adding $\Labs$ to $\LsupAP$.}
    \label{fig:relative_increase}
\end{figure*}

\begin{figure}[ht]
    \centering
    \includegraphics[trim=50 23 50 23,clip,width=\textwidth]{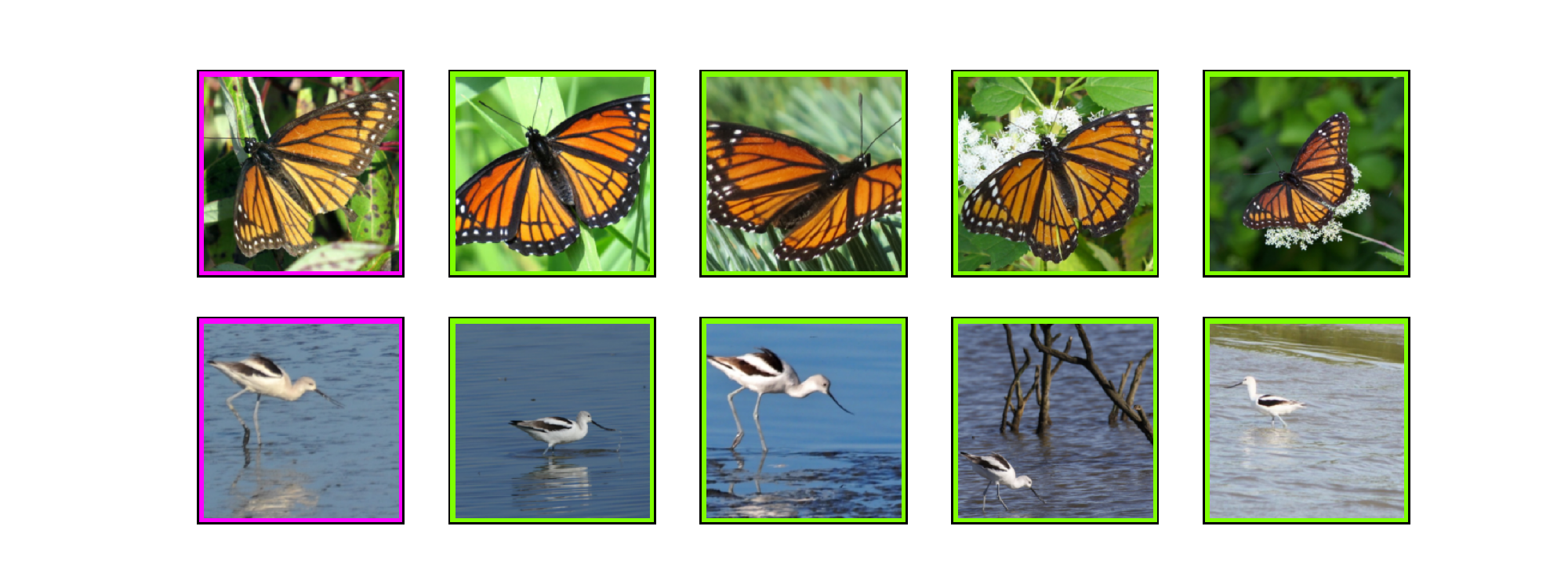}
    \caption{Results on INaturalist: a query (purple) with the 4 most similar retrieved images (green).}
    \label{fig:qualitative_results}
\end{figure}

As a qualitative assessment, we show in \cref{fig:qualitative_results} some results of ROADMAP on INaturalist. We show the queries (in purple) and the 4 most similar retrieved images (in green). We can appreciate the semantic quality of the retrieval. More qualitative results are provided in supplementary Sec. C.

\cref{fig:qualitative_results_compa} shows another qualitative assessment on INaturalist, where ROADMAP corrects some failing cases of the SmoothAP baseline. 

\begin{figure}[ht]
    \centering
    \includegraphics[width=\textwidth]{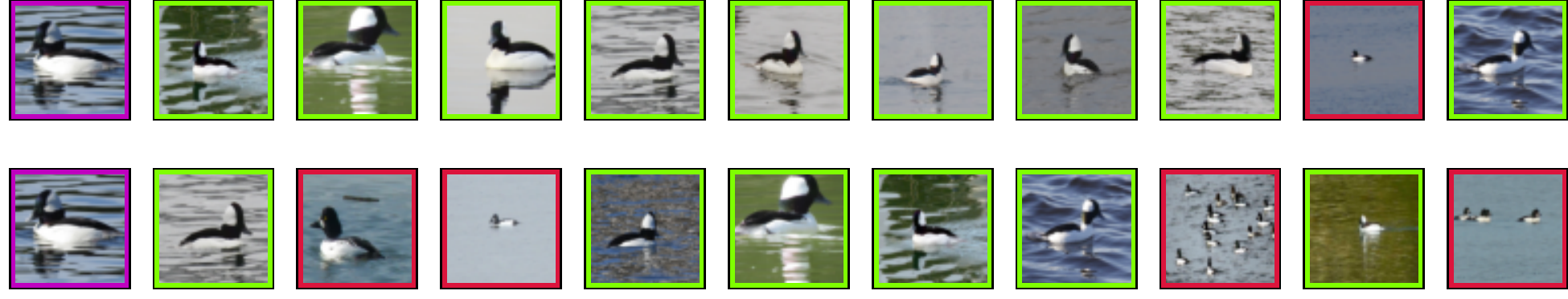}
    \caption{Results on INaturalist: a query (purple) with the 9 most similar retrieved images, green for relevant images, red otherwise. Top line results with ROADMAP. Bottom line results with SmoothAP.}
    \label{fig:qualitative_results_compa}
\end{figure}

\section{Conclusion}
\label{sec:conclusion}

This paper introduces the ROADMAP method for gradient-based optimization of average precision. ROADMAP is based on a smooth rank approximation, leading to the $\LsupAP$ being both accurate and robust. To overcome the lack of decomposability in AP, ROADMAP is equipped with a calibration loss $\Labs$ which aims at reducing the decomposability gap. We provide theoretical guarantees as well as experiments to assess this behavior. 
Experiments show that ROADMAP can combine the strength of ranking methods with the simplicity of a batch strategy.
Without bells and whistles, ROADMAP is able to outperform state-of-the-art performances on three datasets, and remains effective even with small batch sizes.

As any work on image retrieval, our contribution could be applied to critical applications in surveillance scenarios, \eg face recognition or person re-identification. ROADMAP is neither worse nor better than previous work in this regard. Our work is also a data-driven learning method, and thus inherits the risk of perpetuating dataset biases. Future work will focus on improving fair and accurate retrieval by reducing dataset biases. We also plan to relax the need for full supervision to tackle situations more representative to in-the-wild scenarios.

\paragraph*{Acknowledgement} This work was done under a grant from the the AHEAD ANR program (ANR-20-THIA-0002). It was granted access to the HPC resources of IDRIS under the allocation 2021-AD011012645 made by GENCI.

\newpage
\bibliographystyle{plain}
\bibliography{biblio.bib}

\begin{thebibliography}{10}

\bibitem{unifying_mi}
Malik Boudiaf, J{\'e}r{\^o}me Rony, Imtiaz~Masud Ziko, Eric Granger, Marco
  Pedersoli, Pablo Piantanida, and Ismail~Ben Ayed.
\newblock A unifying mutual information view of metric learning: cross-entropy
  vs. pairwise losses.
\newblock In {\em European Conference on Computer Vision}, pages 548--564.
  Springer, 2020.

\bibitem{smoothap}
Andrew Brown, Weidi Xie, Vicky Kalogeiton, and Andrew Zisserman.
\newblock Smooth-ap: Smoothing the path towards large-scale image retrieval.
\newblock In {\em European Conference on Computer Vision}, pages 677--694.
  Springer, 2020.

\bibitem{fastap}
Fatih Cakir, Kun He, Xide Xia, Brian Kulis, and Stan Sclaroff.
\newblock Deep metric learning to rank.
\newblock In {\em Proceedings of the IEEE/CVF Conference on Computer Vision and
  Pattern Recognition}, pages 1861--1870, 2019.

\bibitem{DBLP:conf/sigir/CarvalhoCPSTC18}
Micael Carvalho, R{\'{e}}mi Cad{\`{e}}ne, David Picard, Laure Soulier, Nicolas
  Thome, and Matthieu Cord.
\newblock Cross-modal retrieval in the cooking context: Learning semantic
  text-image embeddings.
\newblock In Kevyn Collins{-}Thompson, Qiaozhu Mei, Brian~D. Davison, Yiqun
  Liu, and Emine Yilmaz, editors, {\em The 41st International {ACM} {SIGIR}
  Conference on Research {\&} Development in Information Retrieval, {SIGIR}
  2018, Ann Arbor, MI, USA, July 08-12, 2018}, pages 35--44. {ACM}, 2018.

\bibitem{vit}
Alexey Dosovitskiy, Lucas Beyer, Alexander Kolesnikov, Dirk Weissenborn,
  Xiaohua Zhai, Thomas Unterthiner, Mostafa Dehghani, Matthias Minderer, Georg
  Heigold, Sylvain Gelly, et~al.
\newblock An image is worth 16x16 words: Transformers for image recognition at
  scale.
\newblock {\em arXiv preprint arXiv:2010.11929}, 2020.

\bibitem{Durand19}
Thibaut Durand, Nicolas Thome, and Matthieu Cord.
\newblock Exploiting negative evidence for deep latent structured models.
\newblock {\em IEEE Transactions on Pattern Analysis and Machine Intelligence},
  41(2):337--351, 2019.

\bibitem{transformer_ir}
Alaaeldin El-Nouby, Natalia Neverova, Ivan Laptev, and Herv{\'e} J{\'e}gou.
\newblock Training vision transformers for image retrieval.
\newblock {\em arXiv preprint arXiv:2102.05644}, 2021.

\bibitem{Engilberge_2019_CVPR}
Martin Engilberge, Louis Chevallier, Patrick Perez, and Matthieu Cord.
\newblock Sodeep: A sorting deep net to learn ranking loss surrogates.
\newblock In {\em Proceedings of the IEEE/CVF Conference on Computer Vision and
  Pattern Recognition (CVPR)}, June 2019.

\bibitem{VSE++}
Fartash Faghri, David~J. Fleet, Jamie~Ryan Kiros, and Sanja Fidler.
\newblock {VSE++:} improving visual-semantic embeddings with hard negatives.
\newblock In {\em British Machine Vision Conference 2018, {BMVC} 2018,
  Newcastle, UK, September 3-6, 2018}, page~12. {BMVA} Press, 2018.

\bibitem{Ge_2018_ECCV}
Weifeng Ge.
\newblock Deep metric learning with hierarchical triplet loss.
\newblock In {\em Proceedings of the European Conference on Computer Vision
  (ECCV)}, September 2018.

\bibitem{DBLP:journals/ijcv/GordoARL17}
Albert Gordo, Jon Almaz{\'{a}}n, J{\'{e}}r{\^{o}}me Revaud, and Diane Larlus.
\newblock End-to-end learning of deep visual representations for image
  retrieval.
\newblock {\em Int. J. Comput. Vis.}, 124(2):237--254, 2017.

\bibitem{hadsell2006dimensionality}
Raia Hadsell, Sumit Chopra, and Yann LeCun.
\newblock Dimensionality reduction by learning an invariant mapping.
\newblock In {\em 2006 IEEE Computer Society Conference on Computer Vision and
  Pattern Recognition (CVPR'06)}, volume~2, pages 1735--1742. IEEE, 2006.

\bibitem{Harwood_2017_ICCV}
Ben Harwood, Vijay Kumar B~G, Gustavo Carneiro, Ian Reid, and Tom Drummond.
\newblock Smart mining for deep metric learning.
\newblock In {\em Proceedings of the IEEE International Conference on Computer
  Vision (ICCV)}, Oct 2017.

\bibitem{resnet50}
Kaiming He, Xiangyu Zhang, Shaoqing Ren, and Jian Sun.
\newblock Deep residual learning for image recognition. corr abs/1512.03385
  (2015), 2015.

\bibitem{He_2018_CVPR}
Kun He, Fatih Cakir, Sarah~Adel Bargal, and Stan Sclaroff.
\newblock Hashing as tie-aware learning to rank.
\newblock In {\em Proceedings of the IEEE Conference on Computer Vision and
  Pattern Recognition (CVPR)}, June 2018.

\bibitem{He_2018_DOAP}
Kun He, Yan Lu, and Stan Sclaroff.
\newblock Local descriptors optimized for average precision.
\newblock In {\em IEEE Conference on Computer Vision and Pattern Recognition
  (CVPR)}, June 2018.

\bibitem{horde}
Pierre Jacob, David Picard, Aymeric Histace, and Edouard Klein.
\newblock Metric learning with horde: High-order regularizer for deep
  embeddings.
\newblock In {\em Proceedings of the IEEE/CVF International Conference on
  Computer Vision}, pages 6539--6548, 2019.

\bibitem{faiss}
Jeff Johnson, Matthijs Douze, and Herv{\'e} J{\'e}gou.
\newblock Billion-scale similarity search with gpus.
\newblock {\em arXiv preprint arXiv:1702.08734}, 2017.

\bibitem{adam}
Diederik~P Kingma and Jimmy Ba.
\newblock Adam: A method for stochastic optimization.
\newblock {\em arXiv preprint arXiv:1412.6980}, 2014.

\bibitem{DBLP:journals/ijcv/LawTC17}
Marc~T. Law, Nicolas Thome, and Matthieu Cord.
\newblock Learning a distance metric from relative comparisons between
  quadruplets of images.
\newblock {\em Int. J. Comput. Vis.}, 121(1):65--94, 2017.

\bibitem{adamw}
Ilya Loshchilov and Frank Hutter.
\newblock Decoupled weight decay regularization.
\newblock {\em arXiv preprint arXiv:1711.05101}, 2017.

\bibitem{DBLP:conf/iccv/ManmathaWSK17}
R.~Manmatha, Chao{-}Yuan Wu, Alexander~J. Smola, and Philipp
  Kr{\"{a}}henb{\"{u}}hl.
\newblock Sampling matters in deep embedding learning.
\newblock In {\em {IEEE} International Conference on Computer Vision, {ICCV}
  2017, Venice, Italy, October 22-29, 2017}, pages 2859--2867. {IEEE} Computer
  Society, 2017.

\bibitem{Mcfee10metriclearning}
Brian Mcfee and Gert Lanckriet.
\newblock Metric learning to rank.
\newblock In {\em In Proceedings of the 27th annual International Conference on
  Machine Learning (ICML}, 2010.

\bibitem{Mohapatra_2018_CVPR}
Pritish Mohapatra, Michal Rolínek, C.V. Jawahar, Vladimir Kolmogorov, and
  M.~Pawan Kumar.
\newblock Efficient optimization for rank-based loss functions.
\newblock In {\em Proceedings of the IEEE Conference on Computer Vision and
  Pattern Recognition (CVPR)}, June 2018.

\bibitem{proxynca}
Yair Movshovitz-Attias, Alexander Toshev, Thomas~K Leung, Sergey Ioffe, and
  Saurabh Singh.
\newblock No fuss distance metric learning using proxies.
\newblock In {\em Proceedings of the IEEE International Conference on Computer
  Vision}, pages 360--368, 2017.

\bibitem{musgrave2020metric}
Kevin Musgrave, Serge Belongie, and Ser-Nam Lim.
\newblock A metric learning reality check.
\newblock In {\em European Conference on Computer Vision}, pages 681--699.
  Springer, 2020.

\bibitem{PML}
Kevin Musgrave, Serge Belongie, and Ser-Nam Lim.
\newblock Pytorch metric learning, 2020.

\bibitem{blackbox}
Marin~Vlastelica P., Anselm Paulus, V{\'{\i}}t Musil, Georg Martius, and Michal
  Rol{\'{\i}}nek.
\newblock Differentiation of blackbox combinatorial solvers.
\newblock In {\em ICLR}, 2020.

\bibitem{pytorch}
Adam Paszke, Sam Gross, Francisco Massa, Adam Lerer, James Bradbury, Gregory
  Chanan, Trevor Killeen, Zeming Lin, Natalia Gimelshein, Luca Antiga, Alban
  Desmaison, Andreas Kopf, Edward Yang, Zachary DeVito, Martin Raison, Alykhan
  Tejani, Sasank Chilamkurthy, Benoit Steiner, Lu~Fang, Junjie Bai, and Soumith
  Chintala.
\newblock Pytorch: An imperative style, high-performance deep learning library.
\newblock In H.~Wallach, H.~Larochelle, A.~Beygelzimer, F.~d\textquotesingle
  Alch\'{e}-Buc, E.~Fox, and R.~Garnett, editors, {\em Advances in Neural
  Information Processing Systems 32}, pages 8024--8035. Curran Associates,
  Inc., 2019.

\bibitem{Radenovic-CVPR18}
Filip Radenovi\'{c}, Ahmet Iscen, Giorgos Tolias, Yannis Avrithis, and
  Ond{\v{r}}ej Chum.
\newblock Revisiting oxford and paris: Large-scale image retrieval
  benchmarking.
\newblock In {\em CVPR}, 2018.

\bibitem{DBLP:conf/eccv/RadenovicTC16}
Filip Radenovic, Giorgos Tolias, and Ondrej Chum.
\newblock {CNN} image retrieval learns from bow: Unsupervised fine-tuning with
  hard examples.
\newblock In Bastian Leibe, Jiri Matas, Nicu Sebe, and Max Welling, editors,
  {\em Computer Vision - {ECCV} 2016 - 14th European Conference, Amsterdam, The
  Netherlands, October 11-14, 2016, Proceedings, Part {I}}, volume 9905 of {\em
  Lecture Notes in Computer Science}, pages 3--20. Springer, 2016.

\bibitem{naverap}
Jerome Revaud, Jon Almaz{\'a}n, Rafael~S Rezende, and Cesar Roberto~de Souza.
\newblock Learning with average precision: Training image retrieval with a
  listwise loss.
\newblock In {\em Proceedings of the IEEE/CVF International Conference on
  Computer Vision}, pages 5107--5116, 2019.

\bibitem{blackboxap}
Michal Rol{\'\i}nek, V{\'\i}t Musil, Anselm Paulus, Marin Vlastelica, Claudio
  Michaelis, and Georg Martius.
\newblock Optimizing rank-based metrics with blackbox differentiation.
\newblock In {\em Proceedings of the IEEE/CVF Conference on Computer Vision and
  Pattern Recognition}, pages 7620--7630, 2020.

\bibitem{mic}
Karsten Roth, Biagio Brattoli, and Bjorn Ommer.
\newblock Mic: Mining interclass characteristics for improved metric learning.
\newblock In {\em Proceedings of the IEEE/CVF International Conference on
  Computer Vision}, pages 8000--8009, 2019.

\bibitem{NIPS2016_6b180037}
Kihyuk Sohn.
\newblock Improved deep metric learning with multi-class n-pair loss objective.
\newblock In D.~Lee, M.~Sugiyama, U.~Luxburg, I.~Guyon, and R.~Garnett,
  editors, {\em Advances in Neural Information Processing Systems}, volume~29.
  Curran Associates, Inc., 2016.

\bibitem{SOP}
Hyun~Oh Song, Yu~Xiang, Stefanie Jegelka, and Silvio Savarese.
\newblock Deep metric learning via lifted structured feature embedding.
\newblock In {\em IEEE Conference on Computer Vision and Pattern Recognition
  (CVPR)}, 2016.

\bibitem{Suh_2019_CVPR}
Yumin Suh, Bohyung Han, Wonsik Kim, and Kyoung~Mu Lee.
\newblock Stochastic class-based hard example mining for deep metric learning.
\newblock In {\em Proceedings of the IEEE/CVF Conference on Computer Vision and
  Pattern Recognition (CVPR)}, June 2019.

\bibitem{proxynca++}
Eu~Wern Teh, Terrance DeVries, and Graham~W Taylor.
\newblock Proxynca++: Revisiting and revitalizing proxy neighborhood component
  analysis.
\newblock In {\em European Conference on Computer Vision (ECCV)}. Springer,
  2020.

\bibitem{deit}
Hugo Touvron, Matthieu Cord, Matthijs Douze, Francisco Massa, Alexandre
  Sablayrolles, and Herv{\'e} J{\'e}gou.
\newblock Training data-efficient image transformers \& distillation through
  attention.
\newblock {\em arXiv preprint arXiv:2012.12877}, 2020.

\bibitem{histogram_loss}
Evgeniya Ustinova and Victor Lempitsky.
\newblock Learning deep embeddings with histogram loss.
\newblock In D.~Lee, M.~Sugiyama, U.~Luxburg, I.~Guyon, and R.~Garnett,
  editors, {\em Advances in Neural Information Processing Systems}, volume~29.
  Curran Associates, Inc., 2016.

\bibitem{inaturalist}
Grant Van~Horn, Oisin Mac~Aodha, Yang Song, Yin Cui, Chen Sun, Alex Shepard,
  Hartwig Adam, Pietro Perona, and Serge Belongie.
\newblock The inaturalist species classification and detection dataset.
\newblock In {\em Proceedings of the IEEE conference on computer vision and
  pattern recognition}, pages 8769--8778, 2018.

\bibitem{CUB}
C.~Wah, S.~Branson, P.~Welinder, P.~Perona, and S.~Belongie.
\newblock {The Caltech-UCSD Birds-200-2011 Dataset}.
\newblock Technical Report CNS-TR-2011-001, California Institute of Technology,
  2011.

\bibitem{multi_similarity}
Xun Wang, Xintong Han, Weilin Huang, Dengke Dong, and Matthew~R Scott.
\newblock Multi-similarity loss with general pair weighting for deep metric
  learning.
\newblock In {\em Proceedings of the IEEE/CVF Conference on Computer Vision and
  Pattern Recognition}, pages 5022--5030, 2019.

\bibitem{xbm}
Xun Wang, Haozhi Zhang, Weilin Huang, and Matthew~R Scott.
\newblock Cross-batch memory for embedding learning.
\newblock In {\em Proceedings of the IEEE/CVF Conference on Computer Vision and
  Pattern Recognition}, pages 6388--6397, 2020.

\bibitem{timm}
Ross Wightman.
\newblock Pytorch image models.
\newblock \url{https://github.com/rwightman/pytorch-image-models}, 2019.

\bibitem{sampling_matters}
Chao-Yuan Wu, R~Manmatha, Alexander~J Smola, and Philipp Krahenbuhl.
\newblock Sampling matters in deep embedding learning.
\newblock In {\em Proceedings of the IEEE International Conference on Computer
  Vision}, pages 2840--2848, 2017.

\bibitem{NIPS2002_c3e4035a}
Eric Xing, Michael Jordan, Stuart~J Russell, and Andrew Ng.
\newblock Distance metric learning with application to clustering with
  side-information.
\newblock In S.~Becker, S.~Thrun, and K.~Obermayer, editors, {\em Advances in
  Neural Information Processing Systems}, volume~15. MIT Press, 2003.

\bibitem{xuan2020hard}
Hong Xuan, Abby Stylianou, Xiaotong Liu, and Robert Pless.
\newblock Hard negative examples are hard, but useful.
\newblock In {\em European Conference on Computer Vision}, pages 126--142.
  Springer, 2020.

\bibitem{hydra}
Omry Yadan.
\newblock Hydra - a framework for elegantly configuring complex applications.
\newblock Github, 2019.

\bibitem{Yue:2007}
Yisong Yue, Thomas Finley, Filip Radlinski, and Thorsten Joachims.
\newblock A support vector method for optimizing average precision.
\newblock In {\em Proceedings of the 30th Annual International ACM SIGIR
  Conference on Research and Development in Information Retrieval}, SIGIR '07,
  pages 271--278, New York, NY, USA, 2007. ACM.

\bibitem{norm_softmax}
Andrew Zhai and Hao{-}Yu Wu.
\newblock Making classification competitive for deep metric learning.
\newblock {\em CoRR}, abs/1811.12649, 2018.

\bibitem{spherical_embedding}
Dingyi Zhang, Yingming Li, and Zhongfei Zhang.
\newblock Deep metric learning with spherical embedding.
\newblock In H.~Larochelle, M.~Ranzato, R.~Hadsell, M.~F. Balcan, and H.~Lin,
  editors, {\em Advances in Neural Information Processing Systems}, volume~33,
  pages 18772--18783. Curran Associates, Inc., 2020.

\bibitem{fewer_is_more}
Yuehua Zhu, Muli Yang, Cheng Deng, and Wei Liu.
\newblock Fewer is more: A deep graph metric learning perspective using fewer
  proxies.
\newblock In H.~Larochelle, M.~Ranzato, R.~Hadsell, M.~F. Balcan, and H.~Lin,
  editors, {\em Advances in Neural Information Processing Systems}, volume~33,
  pages 17792--17803. Curran Associates, Inc., 2020.

\end{thebibliography}

\newpage
\setcounter{section}{0}
\renewcommand\thesection{\Alph{section}}

\section{ROADMAP model}
\label{sec:sup_roadmap}

\subsection{Properties of SupAP \& comparison to SmoothAP}

We further discuss and give additional explanations of the property of our $\LsupAP$ loss function, and especially its comparison with respect to the SmoothAP \cite{smoothap} baseline. 

As shown in Fig. 1.a of the main paper, and discussed in Section 3.1 ("Comparison to SmoothAP"), 
the smooth rank approximation in \cite{smoothap} has several drawbacks, that we show below: 
 \begin{figure*}[!hb]
     \centering
     \includegraphics[scale=.55]{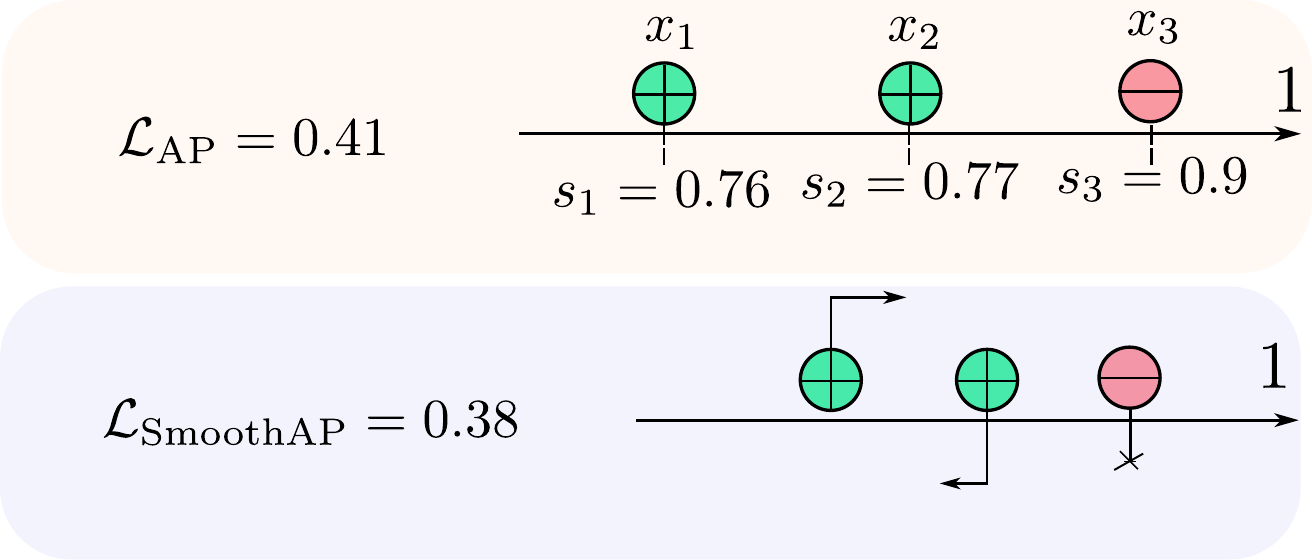}
     \caption{Limitation of the smooth rank approximation in~\cite{smoothap}: contradictory gradient flow for the positives samples $\boldsymbol{x_1}$ and $\boldsymbol{x_2}$ (in green), vanishing gradient for the negative example $\boldsymbol{x_3}$ (in red), and no guarantees of having an upper bound of $\LAP$.}
     \label{fig:comparison_smoothap_sup}
 \end{figure*}

Specifically, we explain in more detail the following three limitations identified in the main paper for SmoothAP \cite{smoothap}, 
~which comes from the use of the sigmoid function to approximate the Heaviside (step) function for computing the rank:

\begin{enumerate}[i]
    
    \item \textbf{Contradictory gradient flow for positives samples:} Firstly we can see on the toy dataset of \cref{fig:comparison_smoothap_sup}
    ~that the gradients of the two positive examples (in green) with SmoothAP have opposite directions. The positive with the lowest rank $\boldsymbol{x_1}$ has a gradient in the good direction, since it leads to increase $\boldsymbol{x_1}$'s score because the correct ordering is not reached (the negative instance $\boldsymbol{x_3}$ has a better rank). But the gradient of the positive with the highest rank $\boldsymbol{x_2}$ is on the wrong direction, since it tends to decrease $\boldsymbol{x_2}$'s score. This is an undesirable behaviour, which comes from the use of the sigmoid in $\LSmoothAP$. In the example of \cref{fig:comparison_smoothap_sup}, we can actually show that

    \begin{equation*}
        \boxed{
        {\frac{\partial \LSmoothAP}{\partial s_1} = - \frac{\partial \LSmoothAP}{\partial s_2}}
        }
    \end{equation*}

    To see this we write :
\begin{align*}
    \frac{\partial \LSmoothAP}{\partial s_1} = &\frac{\partial \LSmoothAP}{\partial \rank^+(x_1)}\cdot\frac{\partial \rank^+(x_1)}{\partial s_1} + \frac{\partial \LSmoothAP}{\partial \rank^+(x_2)} \cdot \frac{\partial \rank^+(x_2)}{\partial s_1} \\ &+ \frac{\partial \LSmoothAP}{\partial \rank^-(x_1)}\cdot\frac{\partial \rank^-(x_1)}{\partial s_1} + \frac{\partial \LSmoothAP}{\partial \rank^-(x_2)}\cdot\frac{\partial \rank^-(x_2)}{\partial s_1} 
\end{align*}

Because $\rank^-(x_2) = \sigma(\frac{s_3-s_2}{\tau})$, we have $\frac{\partial \rank^-(x_2)}{\partial s_1}=0$ and $\frac{\partial \rank^-(x_1)}{\partial s_1}=0$ in the example of \cref{fig:comparison_smoothap_sup}, because ${\rank^-(x_1) = \sigma(\frac{s_3-s_1}{\tau})}$ and ${s_3-s_1}$ falls into the saturation regime of the sigmoid. We get a similar result for the derivative of $\LSmoothAP$ wrt. $s_2$ :
    
\begin{equation*}
 \frac{\partial \LSmoothAP}{\partial s_2} = \frac{\partial \LSmoothAP}{\partial \rank^+(x_1)}\cdot\frac{\partial \rank^+(x_1)}{\partial s_2} + \frac{\partial \LSmoothAP}{\partial \rank^+(x_2)}\cdot\frac{\partial \rank^+(x_2)}{\partial s_2}
\end{equation*}

Furthermore we have :

\begin{equation*}
    \frac{\partial \rank^+(x_1)}{\partial s_1} = -\frac{\partial \rank^+(x_1)}{\partial s_2}
\end{equation*}

Indeed ${\rank^+(x_1) = 1 + \sigma(\frac{s_2-s_1}{\tau})}$, such that ${\frac{\partial \rank^+(x_1)}{\partial s_1} = -\tau\cdot \sigma(\frac{s_2-s_1}{\tau})\left(1-\sigma(\frac{s_2-s_1}{\tau})\right)}$ and ${\frac{\partial \rank^+(x_1)}{\partial s_2} = \tau\cdot\sigma(\frac{s_2-s_1}{\tau})\left(1-\sigma(\frac{s_2-s_1}{\tau})\right)}$. Similarly the derivatives of $\rank^+(x_2)$ wrt. $s_1$ and $s_2$ also have opposite signs: ${\frac{\partial \rank^+(x_2)}{\partial s_1} = -\frac{\partial \rank^+(x_2)}{\partial s_2}}$. It concludes the proof that ${\frac{\partial \LSmoothAP}{\partial s_1} = - \frac{\partial \LSmoothAP}{\partial s_2}}$.

    \item \textbf{Vanishing gradients:} Secondly, SmoothAP \cite{smoothap} has vanishing gradients due to its use of the sigmoid function. This is illustrated on the toy dataset in \cref{fig:comparison_smoothap_sup}. The negative instance $\boldsymbol{x_3}$ has a high score $s_3$, but does not receive any gradient, which does not enable it to lower its score although it would improve the overall ranking. This is because the score difference between $\boldsymbol{x_3}$ and $\boldsymbol{x_2}$ is large, \ie $s_3-s_2=0.13$. Similarly, $s_3-s_1=0.14$. Consequently, both $s_3-s_2$ and $s_3-s_1$ fall into the saturation regime of the sigmoid, preventing to propagate any gradient (see Fig. 3c. in the main paper).

    \item  \textbf{Finally, $\boldsymbol{\LSmoothAP}$ is not an upper bound of $\boldsymbol{\LAP}$}. The use of the sigmoid means that both $\rank^+$ and $\rank^-$ can be over or under estimated. If $\rank^+$ is overestimated (resp. underestimated) $\LSmoothAP$ underestimates $\LAP$ (resp. overestimates). And if $\rank^-$ is overestimated (resp. underestimated) $\LSmoothAP$ overestimates $\LAP$ (resp. overestimated). Therefore, $\LSmoothAP$ can be larger or lower than $\LAP$ in general. In the example of \cref{fig:comparison_smoothap_sup}, we show that $\LSmoothAP$ is lower than $\LAP$.
    
\end{enumerate}

\begin{figure*}[t]
    \centering
    \begin{subfigure}[t]{0.47\textwidth}
        \centering
        \includegraphics[scale=.46]{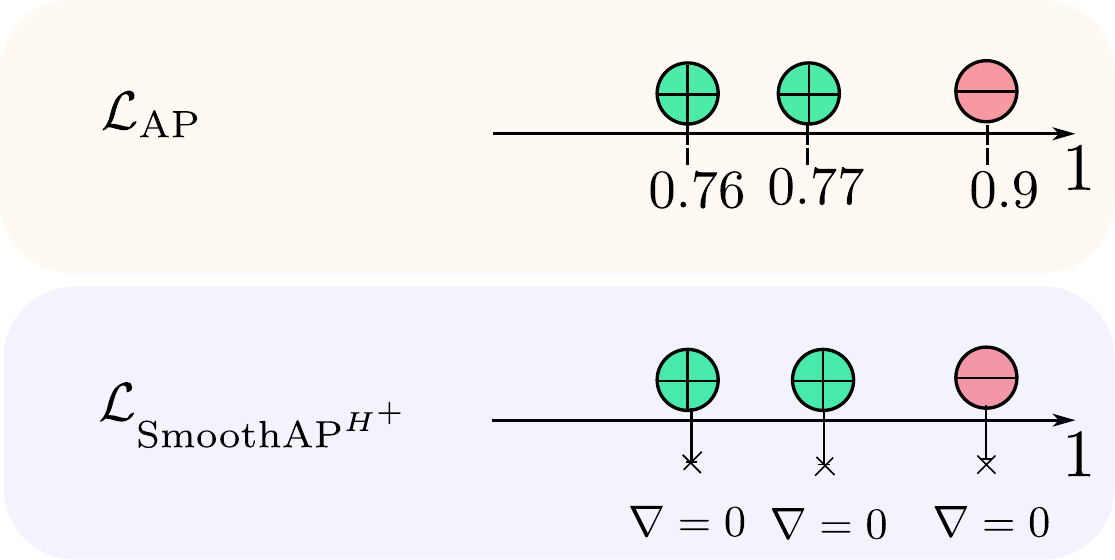}
        \caption{When replacing $H^+$ by the Heaviside function in SmoothAP we stop the unexpected behaviour of the gradient flow. However there is still vanishing gradients.}
        \label{fig:step_smoothap_sup}
    \end{subfigure}
    ~ 
    \begin{subfigure}[t]{0.47\textwidth}
        \centering
        \includegraphics[scale=.46]{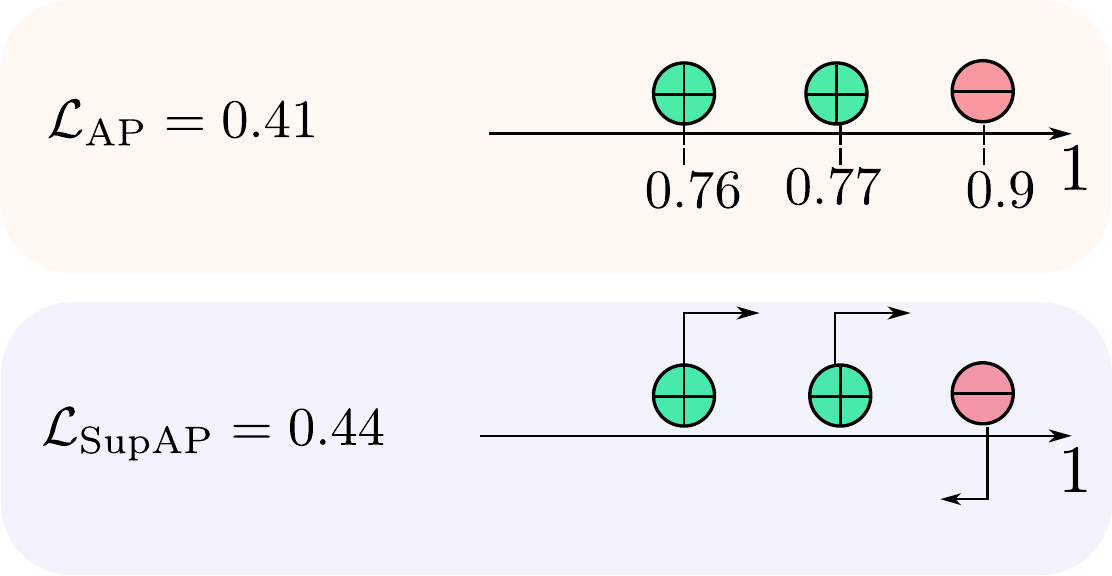}
        \caption{Our $\LsupAP$ has gradients that do not stop until the correct ranking is achieved.}
        \label{fig:upper_smoothap}
    \end{subfigure}
    \caption{We illustrates the different steps to built $\LsupAP$. On \cref{fig:step_smoothap_sup} we change $H^+$ to be the true Heaviside (step) function. On \cref{fig:upper_smoothap} we replace the sigmoid by $H^-$ defined in Eq. (4) of the main paper. Using $H^+$ and $H^-$, $\LsupAP$ is an upper bound of $\LAP$.}
    \label{fig:introb_sup}
\end{figure*}

\textbf{\underline{We address those three issues with $\boldsymbol{\LsupAP}$:}}
\begin{enumerate}[i]
\item \textbf{Using the the true Heaviside (step) function $\mathbf{H^+}$ for $\boldsymbol{\rank^+}$} allows to have the expected behaviour regarding the gradients of positives. When Changing $\mathbf{H^+}$ for $\mathbf{\rank^+}$ in \cref{fig:step_smoothap_sup}, we can see that we fix the problem of opposite gradients for the positive examples $\boldsymbol{x_1}$ and $\boldsymbol{x_2}$ - although the gradient is zero.

\item \textbf{Using $\mathbf{H^-}$ for $\boldsymbol{\rank^-}$ overcomes vanishing gradients}.
~By using $\mathbf{H^-}$ in Eq. (4) in submission, we design a linear function for positive $(s_j-s_k)$ values, where $s_j$ (resp. $s_k$) is the score of a negative (resp. positive) example - see Fig. 3b in the main paper. We can see in \cref{fig:upper_smoothap} that this change enables to have gradients in the correct directions for the two positive instances $\boldsymbol{x_1}$ and $\boldsymbol{x_2}$ (tending to increase their scores), and for the negative instance $\boldsymbol{x_3}$ (tending to decrease its score).

\item \textbf{$\boldsymbol{\LsupAP}$ is an upper bound of $\boldsymbol{\LAP}$}. By the proposed design of $\mathbf{H^-}$ in Eq. (4) in submission, we have $\rank_s^-(k) \geq \rank^-(k)$. Since we do not approximate $\rank^+(k)$ by keeping the Heaviside function, it leads to $\frac{\rank^+(k)}{\rank^+(k)+\rank_s^-(k)} \leq \frac{\rank^+(k)}{\rank^+(k)+\rank^-(k)}$, and therefore $\boldsymbol{\LsupAP} \geq \LAP$.
\end{enumerate}

Overall, $\LsupAP$ has all the desired properties : i) A correct gradient flow during training, ii) No vanishing gradients while the correct ranking is not reached, iii) Being an upper bound on the AP loss $\LAP$.

\subsection{Properties of the $\Labs$ loss function}

We remind the reader of the definition of the decomposability gap given in Eq. (6) of the main paper.

\begin{equation*}
    DG_\text{AP}(\boldsymbol{\theta}) =  \frac{1}{K}  \sum_{b=1}^K \AP_i^b(\boldsymbol{\theta}) - \AP_i(\boldsymbol{\theta})
\end{equation*}

\begin{figure*}[ht]
    \centering
    \includegraphics[scale=.5]{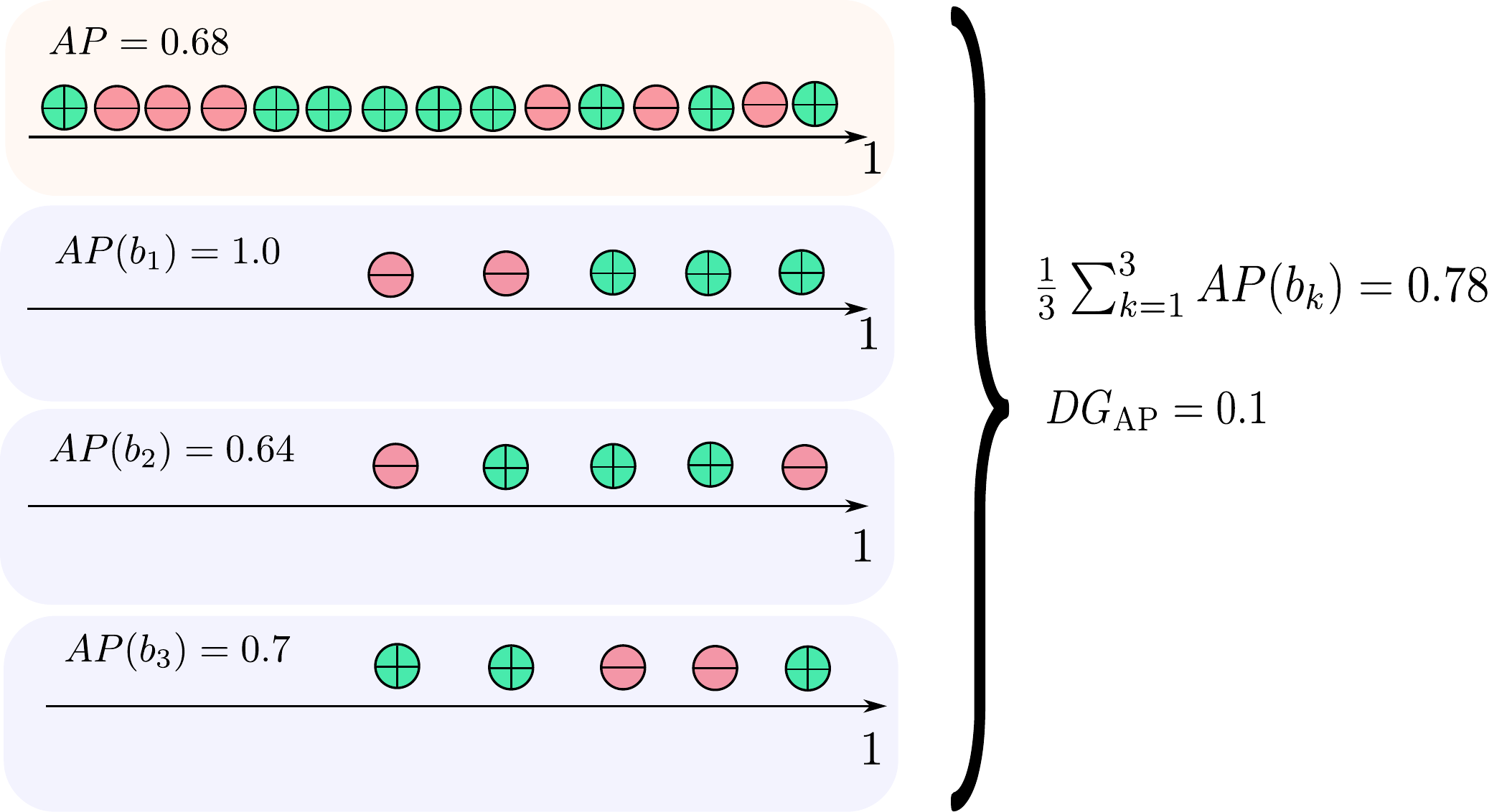}
    \caption{Illustration of the decomposability gap on a toy dataset.}
    \label{fig:dg_ap}
\end{figure*}

We illustrates the decomposability gap, $DG_{AP}$ with the toy dataset of \cref{fig:dg_ap}. The decomposability gap comes from the fact that the AP is not decomposable in mini-batches as we discuss in the Sec. 3.2 of the main paper. The motivation behind $\Labs$ is thus to force the scores of the different batches to aligned as illustrated in the Fig. 2b of the main paper.

\paragraph*{Proof of Eq. (8): Upper bound on the $DG_{\text{AP}}$ with no $\LAP$} We choose a setting for the proof of the upper bound similar to the one used for training, \ie all the batch have the same size, and the number of positive instances per batch (\ie $\mathcal{P}_i^b$) is the same.

\begin{figure*}[t]
     \centering
     \includegraphics[scale=.4]{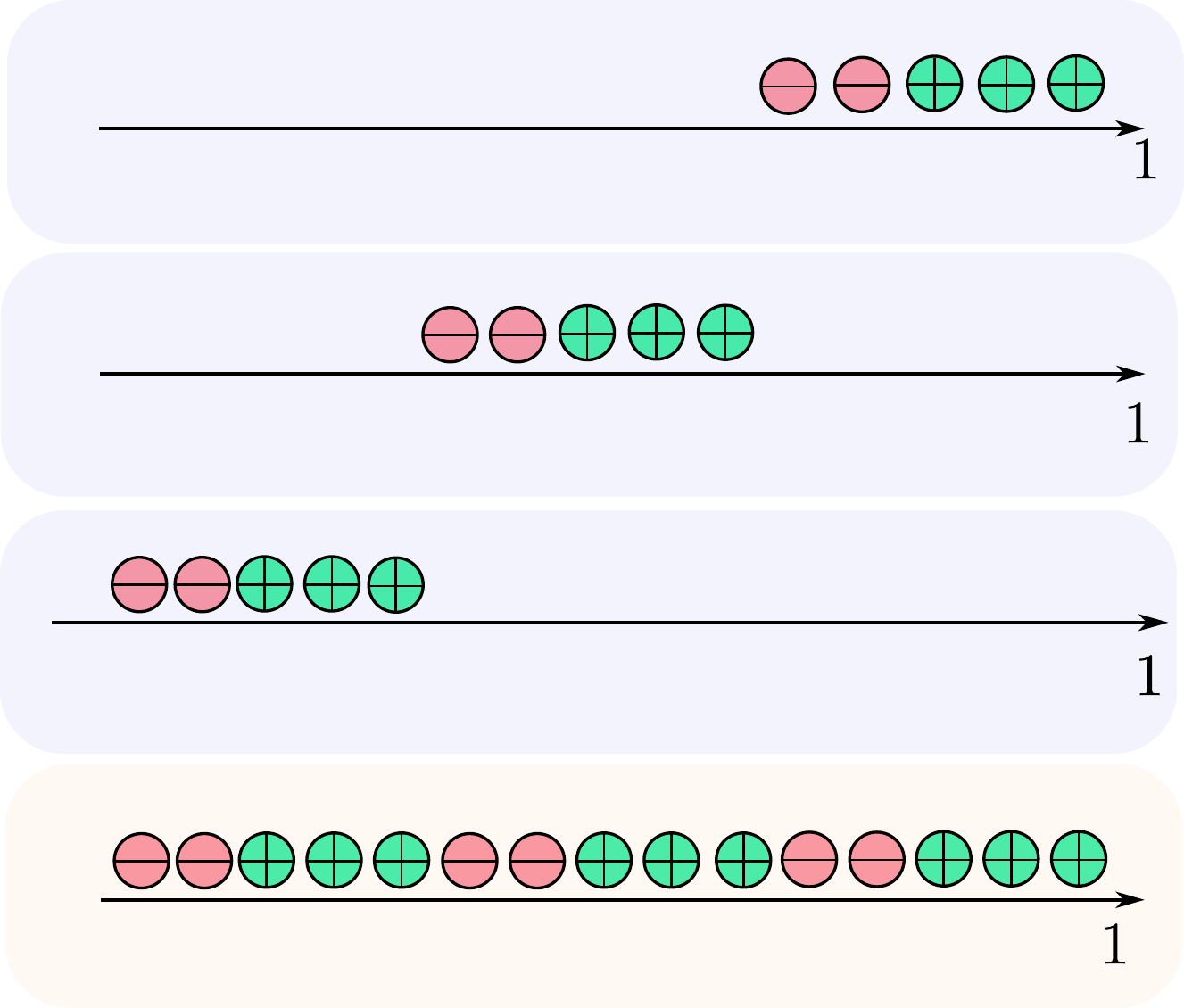}
     \caption{The worst case when computing the global AP would be that each batch is juxtaposed.}
     \label{fig:worst_case_global_ap}
 \end{figure*}

Eq. (8) from the main paper gives an upper bound for $DG_{AP}$. This upper bound is given in the worst case: when the AP has the lowest value guaranteed by the AP on each batch. We illustrate this case in~\cref{fig:worst_case_global_ap}.

In Eq. (8) from the main paper the $1$ in the right hand term comes from the average of AP over all batches:
\begin{equation*}
    \frac{1}{K}\sum_{b=1}^K AP_i^b(\theta)=1
\end{equation*}

We then justify the term in the parenthesis of Eq. (8) in the main paper, which is the lower bound of the AP.
In the global ordering the positive instances are ranked after all the positive instances from previous batches giving the following $\rank^+$: ${j+|\mathcal{P}_i^1|+\dots+|\mathcal{P}_i^{b-1}|}$, with $j$ the $\rank^+$ in the batch, 
~Positive instances are also ranked after all negative instances from previous batches giving $\rank^-$: ${|\mathcal{N}_i^1|+\dots+|\mathcal{N}_i^{b-1}|}$.

Therefore we obtain the resulting upper bound of Eq. (8) of the main paper:

\begin{equation*}
    0 \leq DG_\text{AP} \leq 
1 - \frac{1}{\sum_{b=1}^K |\mathcal{P}_i^b|}\left( \sum_{b=1}^K \sum_{j=1}^B \frac{j + |\mathcal{P}_i^1| + \dots + |\mathcal{P}_i^{b-1}|}{j + |\mathcal{P}_i^1| + \dots + |\mathcal{P}_i^{b-1}| + |\mathcal{N}_i^1| + \dots + |\mathcal{N}_i^{b-1}|} \right ) 
\end{equation*}

\paragraph*{Proof of Eq. (9): Upper bound on the $DG_{AP}$ with $\LAP$} In the main paper we refine the upper bound on $DG_{AP}$ in Eq. (9) by adding $\Labs$ which calibrates the absolute scores across the mini-batches.

We now write that each positive instance that respects the constraint of $\Labs$ is ranked after the positive instances of previous batch that respect the constraint giving the following $\rank^+$: ${j+G_1^++\dots+G_{b-1}^+}$, with $j$ the $\rank^+$ in the current batch. Positive instances are also ranked after the negative instances of previous batches that do not respect the constraints yielding $\rank^-$ : ${E_1^-+\dots+E_{b-1}^-}$.

We then write that positive instances that do not respect the constraints are ranked after all positive instances from previous batches and the positive instances respecting the constraints of the current batch giving $\rank^+$ : ${j+G_b^+|\mathcal{P}_i^1|+\dots+|\mathcal{P}_i^{b-1}|}$. They also are ranked after all the negative instances from previous batches giving $\rank^-$ : ${|\mathcal{N}_i^1|+\dots+|\mathcal{N}_i^{b-1}|}$.

Resulting in Eq. (9) from the main paper:

\begin{align*}
    0 \leq DG_\text{AP}  \leq  1 - \frac{1}{\sum_{b=1}^K |\mathcal{P}_i^b|} \Bigg( & \sum_{b=1}^K \bigg[ \sum_{j=1}^{G^+_b} \frac{j + G^+_1 + \dots + G^+_{b-1}}{j + G^+_1 + \dots + G^+_{b-1} + E^-_1 + \dots E^-_{b-1}} + \\
    & \sum_{j=1}^{E^+_b} \frac{j + G^+_{b} + |\mathcal{P}_i^1| + \dots + |\mathcal{P}_i^{b-1}|}{j + G^+_{b} + |\mathcal{P}_i^1| + \dots + |\mathcal{P}_i^{b-1}| + |\mathcal{N}_i^1| + \dots + |\mathcal{N}_i^{b-1}|} \bigg]  \Bigg)
\end{align*}

\subsection{Choice of $\delta$}

\begin{figure*}[t]
    \centering
    \includegraphics[scale=.5]{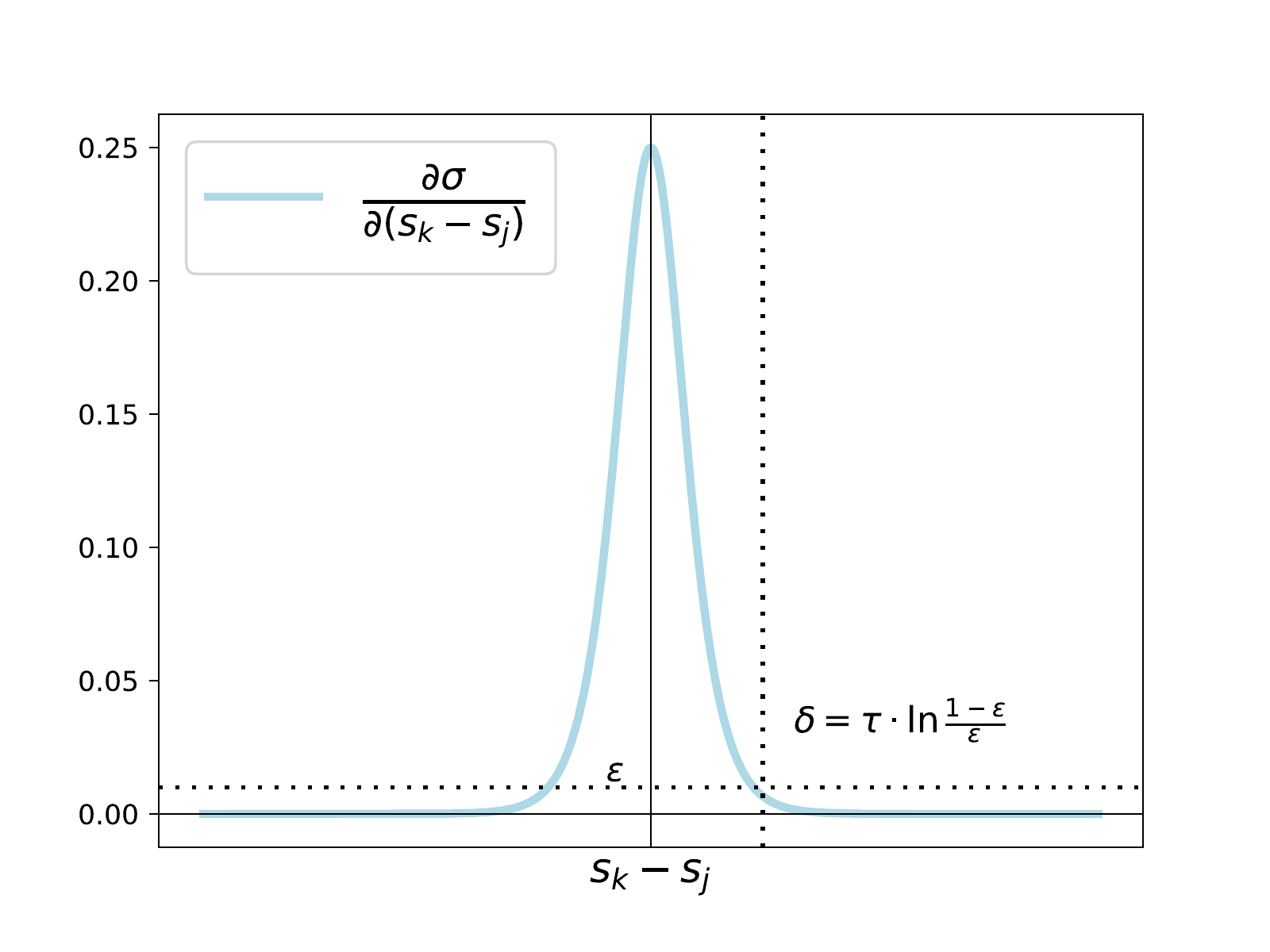}
    \caption{Gradient of the temperature scaled sigmoid ($\tau=0.01$) \vs the difference of scores $s_k-s_j$ of a negative pair.}
    \label{fig:choice_of_delta}
\end{figure*}

In the main paper we introduce $\delta$ in Eq. (4) to define $H^-$. We choose $\delta$ as the point where the gradient of the sigmoid function becomes low $< \epsilon$, and we then have $\delta=\tau\cdot\ln\frac{1-\epsilon}{\epsilon}$. This is illustrated in \cref{fig:choice_of_delta}. For our experiments we use $\epsilon=10^{-2}$ giving $\delta\simeq0.05$.

\section{Experiments}

\subsection{Metrics}

We detail here the performance metrics that we use to evaluate our models.

\paragraph{Recall@K}

The Recall@K metrics (\cref{eq:ratk}) is often used in the literature. For a single query the Recall@K is 1 if a positive instance is in the K nearest neighbors, and 0 otherwise. The Recall@K is then averaged on all the queries. Researcher use different values of K for a given dataset (\eg 1, 2, 4, 8 on CUB).

\begin{equation}
    R@K = \frac{1}{M} \sum_{i=1}^{M} r(i)\text{,~~~where~} r(i) = 
    \begin{cases}
      1 \; \text{if a positive instance has a ranking smaller than i} \\
      0 \quad \text{otherwise}
    \end{cases}
    \label{eq:ratk}
\end{equation}

\paragraph{mAP@R}

Recently, the mAP@R (\cref{eq:mapatr}) has been introduced in \cite{musgrave2020metric}. The authors show that this metric is less noisy and better captures the performance of a model. The mAP@R is a partial AP, computed on the R first instances retrieved, with R being set to the number of positive instances wrt. a query. mAP@R is a lower bound of the AP (mAP@R = AP when the correct ranking is achieved, \ie mAP@R = AP = 1).

\begin{equation}
    mAP@R_i = \frac{1}{R} \sum_{j=1}^{R} P(j)  \text{,~~~where~} P(j) = 
    \begin{cases}
      \text{precision at j} \; \text{if the jth retrieval is correct} \\
      0 \quad \text{otherwise}
    \end{cases}
    \label{eq:mapatr}
\end{equation}

\subsection{Detail on experimental setup}

In this section, we describe the experimental setup used in the Sec. 4.1 of the main paper, and the Sec. B of the supplementary.

We use standard data augmentation strategy during training: images are resized so that their shorter side has a size of 256, we then make a random crop that has a size between 40 and 256, and aspect ratio between 3/4 and 4/3. This crop is then resized to 224x224, and flipped horizontally with a 50\% chance. During evaluation, images are resized to 256 and then center cropped to 224.

We use two different strategy to sample each mini-batch.
On CUB and INaturalist we choose a batch size (\eg 128) and a number of samples per classes (\eg 4). We then randomly sample classes (\eg 32) to construct our batches.
For SOP we use the hard sampling strategy from \cite{fastap}. For each pair of category (\eg bikes and coffee makers) we use the preceding sampling strategy. This sampling techniques is used because it yields harder and more informative batches. The intuition behind this sampling is that it will be harder to discriminate two bikes from one another, than a bike and a sofa.

We train the ResNet-50 models using Adam \cite{adam}.
On CUB we train our models with a learning rate of $10^{-6}$ for 200 epochs.
For SOP and INaturalist we take the same scheduling as in \cite{smoothap}. We set the learning rate for the backbone to $10^{-5}$ and the double for the added linear projection layer. We drop the learning rate by 70\% on the epochs 30 and 70. Finally the models are trained for 100 epochs on SOP and 90 on INaturalist (as in \cite{smoothap}).

We train the DeiT transformers models using AdamW \cite{adamw} as in \cite{transformer_ir}. On INaturalist we use the same schedule as when training ResNet-50, with a learning rate of $10^{-5}$. On SOP we train for 75 epochs with a learning rate of $10^{-5}$ which is dropped by 70\% at epochs 25 and 50. Finally on CUB we train the models for about 100 epochs with a learning rate of $10^{-6}$.

\subsection{Details of the backbones used}

We briefly describe the backbones used throughout out the experiments presented in the main paper and the supplementary.

\paragraph*{ResNet-50 \cite{resnet50}} We use the well-known convolutional neural network ResNet-50. We remove the linear classification layer. We also add a linear projection layer to reduce the dimension (\eg from 2048 to 512).

\paragraph*{DeiT \cite{deit}} Recently transformer models have been introduced for computer vision \cite{vit,deit}. They establish new state-of-the-art performances on computer vision tasks. We use the DeiT-S from \cite{deit} which has less parameters than the ResNet-50 ($\sim$ 21 million for DeiT \vs 25 for ResNet-50). We use the pretrained version with distillation from \cite{deit} and its implementation in the \texttt{timm} library \cite{timm}.

\subsection{ROADMAP validation}

\paragraph*{Comparison to AP approximations} We compare in \cref{tab:compa_ranking_losses_sup} ROADMAP \vs other ranking losses on different settings : a batch size of 128 and two backbones (ResNet-50 and DeiT). We conduct this comparison on 5 runs to show the statistical improvement of our method compared to other ranking losses baselines.

We observe that our method outperforms recent ranking losses on the two backbones and the three datasets. On SOP and CUB, ROADMAP has a high increase for the mAP@R, of +1pt on CUB and +2pt on SOP. The performance improvement is greater on the large scale dataset INaturalist with $\sim$+3.5pt with a ResNet-50 backbone and $\sim$+2pt with a DeiT backbone of mAP@R. This trend is the same as in the comparison of the main paper (Table 1).

\begin{table}[h!]
    \caption{Comparison between ROADMAP and state-of-the-art AP ranking based losses on three image retrieval datasets. \textit{Bck} in the first column stands for bakcbone. Models are trained with a batch size of 128.}
    \setlength\tabcolsep{3pt}
    \label{tab:compa_ranking_losses_sup} 
    \centering
    \begin{tabularx}{\textwidth}{>{\small} l>{\small} l >{\small}c>{\small}c >{\small}c>{\small}c >{\small}c >{\small}c }
        \toprule
         && \multicolumn{2}{c}{CUB} & \multicolumn{2}{c}{SOP} & \multicolumn{2}{c}{INaturalist} \\
         \midrule
         Bck & Method & R@1 & mAP@R & R@1 & mAP@R & R@1 & mAP@R \\
         \midrule
         \multirow{5}{*}{\rotatebox[origin=c]{90}{ResNet-50}}
         & FastAP \cite{fastap} & 61.28$\pm$0.37 & 24.11$\pm$0.16 & 78.97$\pm$0.05 & 52.23$\pm$0.09 & 57.23$\pm$0.05 & 22.17$\pm$0.05 \\
         & SoftBinAP \cite{naverap} & 61.70$\pm$0.10 & 24.29$\pm$0.16 & 80.30$\pm$0.21 & 53.69$\pm$0.27 & 60.88$\pm$0.06 & 23.22$\pm$0.05 \\
         & BlackBoxAP \cite{blackboxap} & 61.96$\pm$0.28 & 23.83$\pm$0.14 & 80.97$\pm$0.07 & 54.49$\pm$0.15 & 59.53$\pm$0.12 & 19.62$\pm$0.02 \\
         & SmoothAP \cite{smoothap} & 62.45$\pm$0.48 & 24.32$\pm$0.1 & 81.13$\pm$0.05 & 54.74$\pm$0.16 & 64.48$\pm$0.05 & 24.33$\pm$0.07 \\
         & ROADMAP (ours) & \textbf{64.05}$\pm$0.51 & \textbf{25.27}$\pm$0.12 & \textbf{82.20}$\pm$ 0.09 & \textbf{56.64}$\pm$0.09 & \textbf{68.15}$\pm$0.10 & \textbf{27.01}$\pm$0.10 \\
         \midrule
         \multirow{5}{*}{\rotatebox[origin=c]{90}{DeiT}}
         & FastAP \cite{fastap} & 73.42$\pm$0.22 & 31.96$\pm$0.06 & 82.92$\pm$0.07 & 59.06$\pm$0.03 & 62.18$\pm$0.07 & 25.48$\pm$0.10 \\
         & SoftBinAP \cite{naverap} & 74.84$\pm$0.11 & 33.57$\pm$0.08 & 84.09$\pm$0.05 & 60.53$\pm$0.07 & 65.97$\pm$0.13 & 27.57$\pm$0.09 \\
         & BlackBoxAP \cite{blackboxap} & 75.45$\pm$0.22 & 33.97$\pm$0.10 & 84.07$\pm$0.09 & 60.20$\pm$0.05 & 70.29$\pm$0.10 & 29.44$\pm$0.06 \\
         & SmoothAP \cite{smoothap} & 76.02$\pm$0.14 & 34.69$\pm$0.08 & 84.28$\pm$0.06 & 60.49$\pm$0.17 & 69.80$\pm$0.08 & 29.56$\pm$0.04 \\
         & ROADMAP (ours) & \textbf{77.14}$\pm$0.12 & \textbf{36.30}$\pm$0.08 & \textbf{85.44}$\pm$ 0.06 & \textbf{62.73}$\pm$0.06 & \textbf{72.81}$\pm$0.11 & \textbf{31.31}$\pm$0.10 \\
         \bottomrule
    \end{tabularx}
\end{table}

We perform a paired student t-test to further asses the statistical significance of the performance boost obtained with ROADMAP. We compute the p-values for both the R@1 and mAP@R:
~it turns out that the p-values are never larger than $0.001$, meaning that the gain is statistically significant (with a risk less than $0.1$\%).

\paragraph*{Ablation studies}

In \cref{tab:supp_ablation_study} we extend the ablation studies of the main paper (Table 2 of main paper) to other settings, including more batch sizes (32, 128, 224, 384) and two backbones (ResNet-50 and DeiT).
On all settings $\LsupAP$ outperforms the $\LSmoothAP$ baseline by almost $\sim$+0.5pt consistently, and almost +1pt on every setting for INaturalist. When we add $\Labs$ the gain is further increased. As noticed in Table 2 (main paper) the gain when adding $\Labs$ is particularly noticeable on the large scale dataset INaturalist with boost in performances that can be up to +3.3pt of mAP@R for the ResNet-50 with a batch size 32.

\begin{table}[t]
    \caption{Ablation study for the impact of our two contribution \vs the SmoothAP baseline for the three datasets and different batch sizes, with a ResNet-50 backbone \cite{resnet50}}
    \setlength\tabcolsep{3pt}
    \label{tab:supp_ablation_study} 
    \begin{tabularx}{\textwidth}{ l l cc |YY|YY|YY  }
        \toprule
        &&&& \multicolumn{2}{c}{CUB} & \multicolumn{2}{c}{SOP} & \multicolumn{2}{c}{INaturalist} \\
        \midrule
         BS & Method & $H^-$ & $\mathcal{L}_{calibr.}$ & R@1 & mAP@R & R@1 & mAP@R & R@1 & mAP@R \\
         \hline
         \multirow{3}{*}{32}
         & SmoothAP &  \xmark & \xmark & 61.84 & 23.76 & 79.96 & 53.21 & 53.25 & 16.4 \\
         & SupAP & \cmark & \xmark  & 62.58 & 24.12 & 80.51 & 53.85 & 55.01 & 17.13 \\
         & ROADMAP &  \cmark & \cmark & \textbf{63.69} & \textbf{24.97} & \textbf{80.74} & \textbf{54.68} & \textbf{56.43} & \textbf{20.43}  \\
        \midrule
        \multirow{3}{*}{128}
         & SmoothAP & \xmark & \xmark & 62.81 & 24.44 & 81.19 & 54.96 & 64.53 & 24.26 \\
         & SupAP & \cmark & \xmark  & 63.18 & 24.9 & 81.72 & 55.65 & 65.79 & 24.77 \\
         & ROADMAP & \cmark & \cmark & \textbf{64.18} & \textbf{25.38} & \textbf{82.18} & \textbf{56.64} & \textbf{68.28} & \textbf{27.13} \\
        \midrule
        \multirow{3}{*}{224}
        & SmoothAP & \xmark & \xmark & 62.93 & 24.69 & 81.2 & 54.73 & 66.62 & 26.08 \\
        & SupAP & \cmark & \xmark  & 64.08 & 25.13 & 81.88 & 55.75 & 67.43 & 26.32 \\
        & ROADMAP & \cmark & \cmark & \textbf{64.65} & \textbf{25.51} & \textbf{82.3} & \textbf{56.55} & \textbf{69.28} & \textbf{27.74} \\
        \midrule
        \multirow{3}{*}{384}
        & SmoothAP & \xmark & \xmark & 63.69 & 24.89 & 81.45 & 55.1 & 67.39 & 26.77 \\
        & SupAP & \cmark & \xmark  & 64.64 & 25.27 & 81.94 & 55.78 & 68.37 & 27.24  \\
        & ROADMAP & \cmark & \cmark & \textbf{64.69} & \textbf{25.36} & \textbf{82.31} & \textbf{56.47} & \textbf{69.19} & \textbf{27.85} \\
         \bottomrule
    \end{tabularx}
\end{table}

In \cref{tab:supp_ablation_study_deit} we extend ablation studies with a transformer backbone (DeiT). We observe the same trend as in \cref{tab:supp_ablation_study}. $\LsupAP$ is consistently better than the $\LSmoothAP$ baseline, with gain up to more than 1pt (\eg on batch size 128 on INaturalist). $\Labs$ further lifts the performances on the three datasets and all batch sizes.

\begin{table}[t]
    \caption{Ablation study for the impact of our two contribution \vs the SmoothAP baseline for the three datasets and different batch sizes, with a DeiT backbone \cite{deit}}
    \setlength\tabcolsep{3pt}
    \label{tab:supp_ablation_study_deit} 
    \begin{tabularx}{\textwidth}{ l l cc |YY|YY|YY  }
        \toprule
        &&&& \multicolumn{2}{c}{CUB} & \multicolumn{2}{c}{SOP} & \multicolumn{2}{c}{INaturalist} \\
         \midrule
         BS &Method & $H^-$ & $\mathcal{L}_{calibr.}$ & R@1 & mAP@R & R@1 & mAP@R & R@1 & mAP@R \\
         \hline
        \multirow{3}{*}{128}
         & SmoothAP &  \xmark & \xmark & 76.2 & 34.7 & 84.16 & 60.18 & 69.83 & 29.49 \\
         & SupAP & \cmark & \xmark  & 76.33 & 34.91 & 84.74 & 61.29 & 71.12 &  30.5 \\
         & ROADMAP & \cmark & \cmark & \textbf{77.09} & \textbf{35.76} & \textbf{85.44} & \textbf{62.57} & \textbf{72.82} & \textbf{31.36} \\
        \midrule
         \multirow{3}{*}{224}
         & SmoothAP &  \xmark & \xmark & 76.38 & 35.33 & 84.3 & 60.49 & 70.55 & 30.25 \\
         & SupAP & \cmark & \xmark  & 76.47 & 35.67 & 84.77 & 61.38 & 71.9 & 31.31  \\
         & ROADMAP & \cmark & \cmark & \textbf{77.14} & \textbf{36.18} & \textbf{85.56} & \textbf{62.75} & \textbf{73.64} & \textbf{31.82} \\
        \midrule
        \multirow{3}{*}{384}
        & SmoothAP & \xmark & \xmark & 76.72 & 35.86 & 84.66 & 61.26 & 71.09 & 30.89 \\
        & SupAP & \cmark & \xmark & 77.13 & 36.17 & 85.01 & 61.76 & 72.55 & 31.89  \\
        & ROADMAP & \cmark & \cmark & \textbf{77.38} & \textbf{36.23} & \textbf{85.35} & \textbf{62.29} & \textbf{73.64} & \textbf{32.12} \\
         \bottomrule
    \end{tabularx}
\end{table}
\paragraph*{Comparison to state of the art method}

We show in \cref{tab:embedding_dim} the impact of increasing the embedding dimension when using ResNet-50. All metrics improve on the three datasets when the embedding dimension increases. We observe a gain particularly important on CUB and SOP with $\sim$+1pt in R@1 and mAP@R.

Choosing an embedding size of 2048 further boost the performances of ROADMAP, yielding competitive performances on CUB and state-of-the-art performances for SOP and INaturalist.

\begin{table}[h!]
    \caption{Difference in performance when using an embedding size of 512 \vs 2048 with a ResNet-50 backbone, on the three datasets. Performances are obtained with the same setup as described in the Sec. 4.2 of the main paper.}
    \label{tab:embedding_dim}
    \centering
    \begin{tabular}{ l c cc cc cc }
        \toprule
         && \multicolumn{2}{c}{CUB} &\multicolumn{2}{c}{SOP} & \multicolumn{2}{c}{INaturalist}\\
         \midrule
         Method & dim & R@1 & mAP@R & R@1 & mAP@R & R@1 & mAP@R \\
         \midrule
         ROADMAP (ours) & 512 & 68.5 & 27.97 & 83.19 & 58.05 & 69.19 & 27.85 \\
         ROADMAP (ours) & 2048 & \textbf{69.87} & \textbf{28.8} & \textbf{83.77} & \textbf{59.38} & \textbf{69.62} & \textbf{27.87} \\
         \bottomrule
    \end{tabular}
\end{table}

\paragraph*{Preliminary results on Landmarks retrieval} We show in \cref{tab:landmark_retrieval} preliminary experiments to evaluate ROADMAP on $\mathcal{R}$Oxford and $\mathcal{R}$Paris~\cite{Radenovic-CVPR18}, by training our model on the SfM-120k dataset and using the standard GitHub code for evaluation\footnote{\url{https://github.com/filipradenovic/cnnimageretrieval-pytorch}}.

We can see that ROADMAP is significantly better than~\cite{transformer_ir} with the DeiT-S~\cite{deit} on $\mathcal{R}$Oxford and $\mathcal{R}$Paris medium protocol, and has similar performances for $\mathcal{R}$Paris hard protocol. This highlights the relevance of using ROADMAP instead of the contrastive loss used in~\cite{transformer_ir}.

\begin{table}[h!]
    \caption{Comparison of ROADMAP vs IRT~\cite{transformer_ir} on $\mathcal{R}$Oxford and $\mathcal{R}$Paris~\cite{Radenovic-CVPR18}. Models are DeiT-S~\cite{deit}, ROADMAP is trained with a batch size of 128.}
    \label{tab:landmark_retrieval}
    \centering
    \begin{tabular}{ l cc cc }
        \toprule
         \multirow{2}{*}{Method}& \multicolumn{2}{c}{$\mathcal{R}$Oxford} &\multicolumn{2}{c}{$\mathcal{R}$Paris}\\
         & Medium & Hard & Medium & Hard \\
         \midrule
         IRT~\cite{transformer_ir} & 34.5 & 15.8 & 65.8 & 42.0 \\
         ROADMAP (ours) & \textbf{38.9} & \textbf{20.7} & \textbf{67.5} & \textbf{42.3} \\
         \bottomrule
    \end{tabular}
\end{table}


\subsection{Model analysis}

\paragraph{Hyperparameters}

\begin{figure*}[h!]
    \centering
    \begin{subfigure}[t]{0.32\textwidth}
        \begin{tikzpicture}[scale=0.5]
        \begin{axis}[
            title={},
            xlabel={},
            ylabel={mAP@R},
            xmin=0, xmax=1,
            ymin=0, ymax=56,
            xtick={0.001,0.01,0.1,1.0},
            ytick={},
            legend pos=south east,
            ymajorgrids=true,
            grid style=dashed,
            xmode=log,
            font=\LARGE,
        ]
        \addplot[
            line width=2,
            color=blue,
            mark=x,
            mark size=5,
            ]
            coordinates {
            (0.001,54.78)(0.01,55.28)(0.05,52.55)(0.1,46.86)(1.0,0.1)
            };
        \end{axis}
        \end{tikzpicture}
      \caption{mAP@R \vs $\tau$ for $\LsupAP$.}
      \label{fig:tau_supap}
    \end{subfigure}
    ~
    \begin{subfigure}[t]{0.32\textwidth}
        \begin{tikzpicture}[scale=0.5]
        \begin{axis}[
            title={},
            xlabel={},
            ylabel={},
            xmin=0, xmax=11000,
            ymin=55, ymax=56,
            xtick={0.1,1.0,10.0,100,1000,10000},
            ytick={},
            legend pos=south east,
            ymajorgrids=true,
            grid style=dashed,
            xmode=log,
            font=\LARGE,
        ]
        \addplot[
            line width=2,
            color=blue,
            mark=x,
            mark size=5,
            ]
            coordinates {
            (0.0,55.1)(0.1,55.17)(1.0,55.28)(10.0,55.37)(100,55.65)(1000,55.84)(10000,55.66)
            };
        \end{axis}
        \end{tikzpicture}
        \caption{mAP@R \vs $\rho$ for $\LsupAP$.}
        \label{fig:rho_supap_sup}
    \end{subfigure}
    
    \caption{Analysis of $\LsupAP$ hyperparameters on SOP (batch size 128).}
    \label{fig:supap_hyperparameters_sup}
\end{figure*}
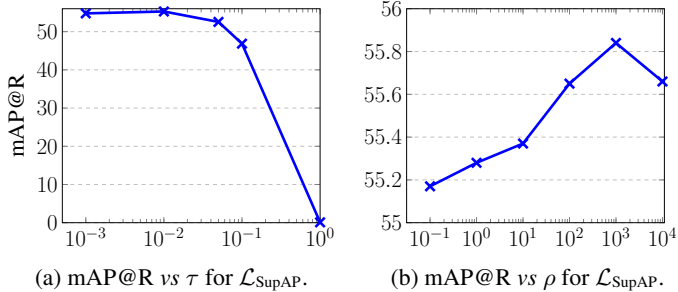

In \cref{fig:supap_hyperparameters_sup} we show the impact of the hyperparameters of $\LsupAP$. We plot the mAP@R \vs $\tau$ in \cref{fig:tau_supap} and mAP@R \vs $\rho$ in \cref{fig:rho_supap_sup}. The experiments are conducted on SOP with a batch size of 128. 

We observe on \cref{fig:tau_supap} that $\LsupAP$ is stable with small values of $\tau$, \ie in the range [0.001, 0.05]. As a reminder we use the default value $\tau=0.01$ in all our results, as it was the suggested value from the SmoothAP paper \cite{smoothap}.

We conduct a study of the impact of $\rho$ in \cref{fig:rho_supap_sup}. We find that $\LsupAP$ is very stable wrt. this hyperparameter. Performances are improving with a greater value of $\rho$ before dropping after $10^4$. The trend follows what was observed in the Fig. 4b of the main paper, although this time using a value if $\rho=10^4$ yields better performances. Using cross-validation to choose an optimal value for $\rho$ may lead to even better performances for $\LsupAP$.

\paragraph{Decomposability gap}

In \cref{tab:dgap_increase} we measure the relative decrease of the decomposability gap $DG_{AP}$ on SOP and CUB test sets. On both datasets we can see that $\Labs$ decreases the decomposability gap.

\begin{table}[h!]
    \caption{Relative decrease of the decomposability gap when adding $\Labs$ to $\LsupAP$ (ROADMAP).}
    \label{tab:dgap_increase}
    \centering
    \begin{tabular}{ c c }
        \toprule
         Dataset &  decrease of $DG_{AP}$ \\
         \midrule
         CUB  & 3.7\% \\
         SOP & 5.4\% \\
         \bottomrule
    \end{tabular}
\end{table}

\subsection{Source code}

We describe in this section the software used for our work, and discuss the computation costs associated with training models presented in this paper.

\paragraph{Librairies} We use several Python libraries often used in image retrieval.

We use \texttt{PyTorch}~\cite{pytorch} as a general framework to implement our neural networks, losses and training loops. We use several utilities from \texttt{PyTorch Metric Learing}~\cite{PML}, an open-source Python library focused on helping researcher working on image retrieval and metric learning. We use \texttt{Faiss}~\cite{faiss} to compute metrics (\ie to perform nearest neighbours search), which is a Python library often used in image retrieval to compute the rankings or the similarity matrix. To load and use the transformer models we use \texttt{timm}~\cite{timm}, a library implementing recent computer vision models, with pretrained weights for most of them. To handle all our config files, we use \texttt{Hydra} \cite{hydra}, this library makes it possible to combine the use of Yaml configuration files and overriding them using the command line.

We use the publicly available implementation of SoftBinAP\footnote{\url{https://github.com/naver/deep-image-retrieval}}~\cite{naverap} which is under a BSD-3 license. The original codes of SmoothAP\footnote{\url{https://github.com/Andrew-Brown1/Smooth_AP}}~\cite{smoothap}, BlackBox\footnote{\url{https://github.com/martius-lab/blackbox-backprop}}~\cite{blackbox,blackboxap} are under an MIT license. For FastAP~\cite{fastap} we use the implementation from \cite{PML} (MIT license), the original implementation of FastAP\footnote{\url{https://github.com/kunhe/FastAP-metric-learning}} is also under an MIT license.

\paragraph*{Compute costs}

We use mixed-precision learning offered within PyTorch \cite{pytorch}. The time and memory consumption are reduced by a factor between $2$ and $3/2$ with no notable difference in performances. We could train all models on 16GiB GPUs, except for models trained with a batch size of 384 which requires a 32GiB GPU.

\textbf{CUB} Models take between 30 minutes and 1 hour to train on a Nvidia Quadro RTX 5000 with 16GiB.

\textbf{SOP} Models take between 4 and 8 hours to train on a Nvidia Quadro RTX 5000 with 16GiB.

\textbf{INaturalist} To train models on INaturalist we were granted access to the IDRIS HPC cluster with Tesla V-100 GPUs (of 16GiB or 32GiB). Models train for approximately 20 hours.

We could not train models with mixed-precision when using BlackBox \cite{blackboxap}. Models trained with it took longer to train (\eg 30 hours on INaturalist) and are more demanding on memory (almost 16GiB with a batch size of 128 while models trained with other loss functions required less than 10Gib).

\section{Qualitative results}

\begin{figure}[ht]
    \centering
    \includegraphics[width=\textwidth]{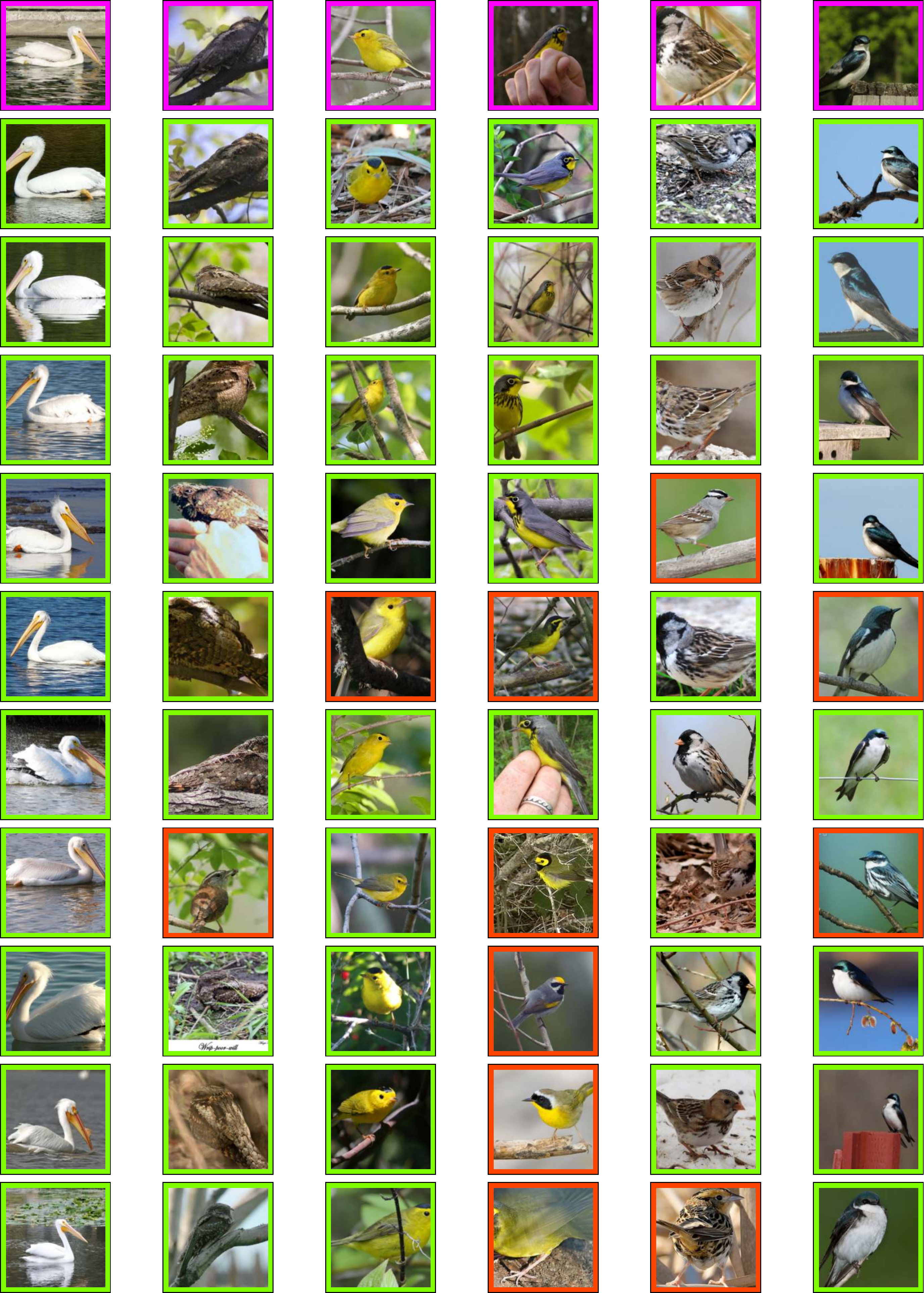}
    \caption{Qualitative results on CUB: a query (purple) with the 10 most similar instances. Relevant (resp. irrelevant) instances are in green (resp. red).}
    \label{fig:qualitative_results_cub}
\end{figure}

\paragraph{CUB} As a qualitative assessment, we show in \cref{fig:qualitative_results_cub} some results of ROADMAP on CUB. We show the queries (in purple) and the 10 most similar retrieved images, with relevant instances in green and irrelevant instances in red.

\begin{figure}[ht]
    \centering
    \includegraphics[width=\textwidth]{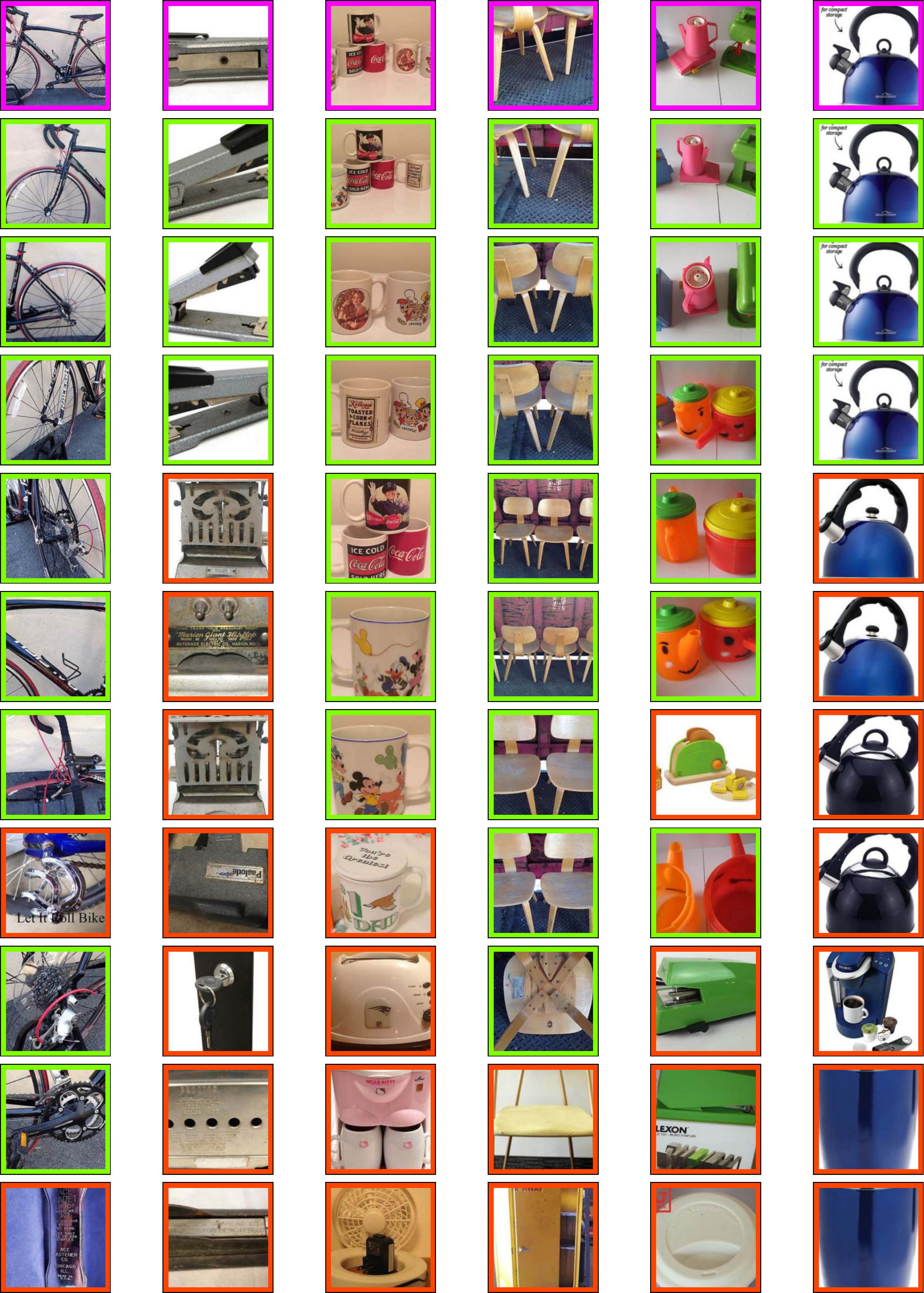}
    \caption{Qualitative results on SOP: a query (purple) with the 10 most similar instances. Relevant (resp. irrelevant) instances are in green (resp. red).}
    \label{fig:qualitative_results_sop}
\end{figure}

\paragraph{SOP} In \cref{fig:qualitative_results_sop} we perform the same assessment for SOP.
In SOP there are fewer relevant instances per query (in average 5). So even for queries that retrieved all the relevant instances, there will be negative instances that have high ranks (in \cref{fig:qualitative_results_sop} ranks that are lower than 10).

\begin{figure}[ht]
    \centering
    \includegraphics[width=\textwidth]{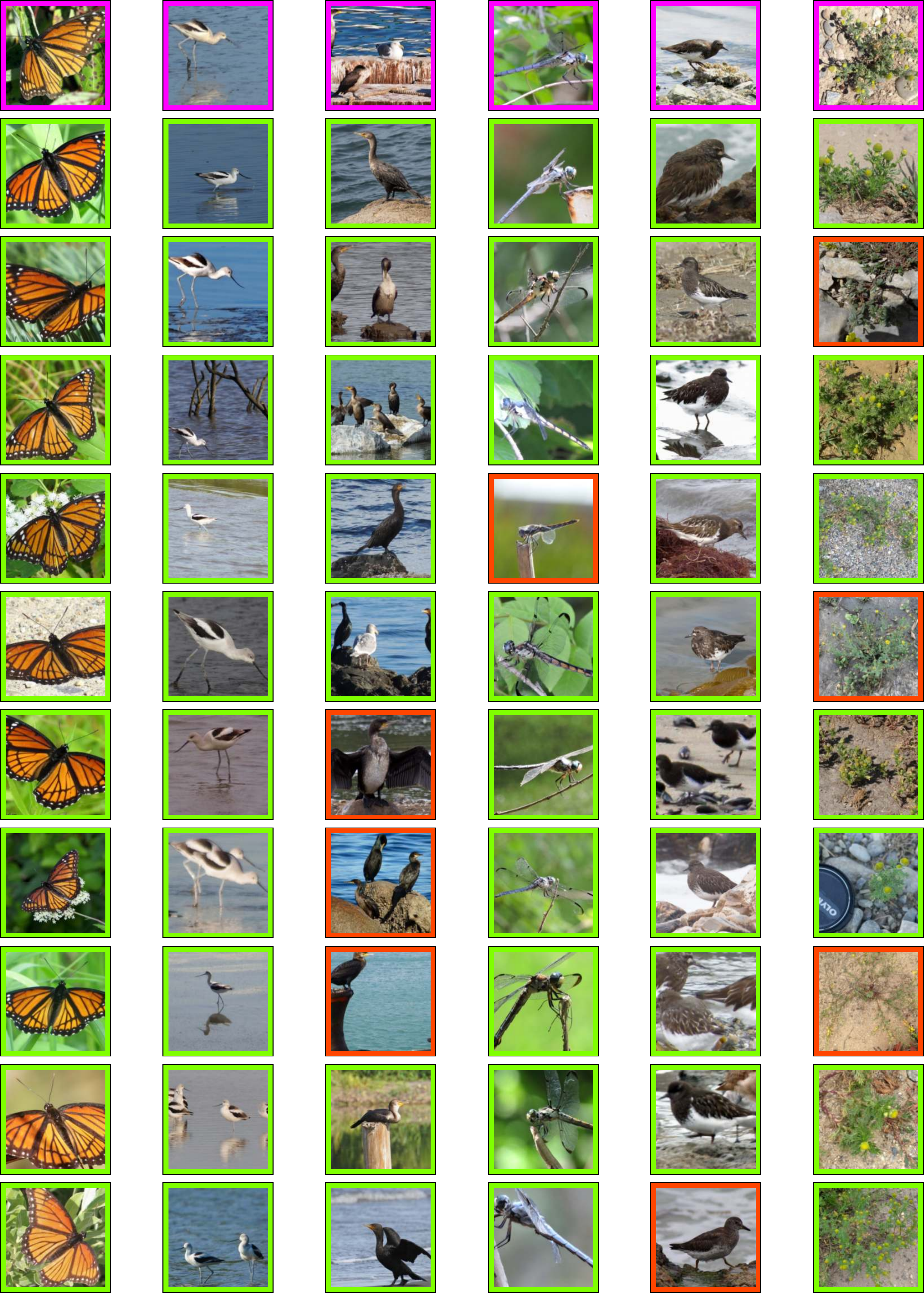}
    \caption{Qualitative results on INaturalist: a query (purple) with the 10 most similar instances. Relevant (resp. irrelevant) instances are in green (resp. red).}
    \label{fig:qualitative_results_inat}
\end{figure}

\paragraph{INaturalist} Finally we show on \cref{fig:qualitative_results_inat} some examples of queries and the 10 most similar instances for a model trained with ROADMAP on INaturalist.

\end{document}